\documentclass[journal]{IEEEtran}
\usepackage{stmaryrd}
\usepackage{bm}
\usepackage{booktabs}
\usepackage{graphicx}
\usepackage{cite}
\usepackage{float}
\usepackage{graphicx}
\usepackage{stfloats}
\usepackage{amsmath}
\usepackage{epstopdf}
\usepackage{multirow}
\usepackage[colorlinks,linkcolor=red,anchorcolor=green,citecolor=green]{hyperref}
\usepackage[numbers,sort&compress]{natbib}

\ifCLASSINFOpdf
\else
\fi

\begin{document}
%
\title{Group Sparsity Residual Constraint for Image Denoising}
%
%
%

\author{Zhiyuan~Zha, Xinggan~Zhang, Qiong~Wang, Lan~Tang and Xin~Liu

\thanks{Z. Zha, X. Zhang and Q. Wang are with the department of Electronic Science and Engineering,
Nanjing University, Nanjing 210023, China. E-mail: zhazhiyuan.mmd@gmail.com.}
\thanks{L. Tang is with the department of Electronic Science and Engineering, Nanjing University,
and National Mobile Commun. Research Lab., Southeast University, Nanjing 210023, China.}
\thanks{L. Xin is with the Center for Machine Vision and Signal Analysis, University of Oulu, 90014, Finland.}
}


\maketitle

\begin{abstract}
Group-based sparse representation has shown great potential in image denoising. However, most existing methods only consider the nonlocal self-similarity (NSS) prior of noisy input image. That is, the similar patches are collected only from degraded input, which makes the quality of image denoising largely depend on the input itself. However, such methods often suffer from a common drawback that the denoising performance may degrade quickly with increasing noise levels. In this paper we propose a new prior model, called group sparsity residual constraint (GSRC). Unlike the conventional group-based sparse representation denoising methods, two kinds of prior, namely, the NSS priors of noisy and pre-filtered images, are used in GSRC. In particular, we integrate these two NSS priors through the mechanism of sparsity residual, and thus, the task of image denoising is converted to the problem of reducing the group sparsity residual. To this end, we first obtain a good estimation of the group sparse coefficients of the original image by pre-filtering, and then the group sparse coefficients of the noisy image are used to approximate this estimation. To improve the accuracy of the nonlocal similar patch selection, an adaptive patch search scheme is designed. Furthermore, to fuse these two NSS prior better, an effective iterative shrinkage algorithm is developed to solve the proposed GSRC model. Experimental results demonstrate that the proposed GSRC modeling outperforms many state-of-the-art denoising methods in terms of the objective and the perceptual metrics.
\end{abstract}

\begin{IEEEkeywords}
Image denoising, group sparsity residual constraint, nonlocal self-similarity, adaptive patch search, iterative shrinkage algorithm.
\end{IEEEkeywords}

\IEEEpeerreviewmaketitle

\section{Introduction}

\IEEEPARstart{A}{s} a classical problem in low level vision, image denoising has been widely studied over the last half century  due to its practical significance. The goal of image denoising is to estimate the clean image $\textbf{\emph{X}}$ from its noisy observation $\textbf{\emph{Y}}=\textbf{\emph{X}} +\textbf{\emph{V}}$, where $\textbf{\emph{V}}$ is additive white Gaussian noise (AWGN). In the past three decades, extensive studies have been conducted on developing various methods for image denoising \cite{1,2,3,4,5,67,6,7,8,64,10,9,11,40,60,35,36,61,63,68}. Due to the ill-posed nature of image denoising,  it has been widely recognized that the prior knowledge of images plays a key role in enhancing the performance of image denoising methods. A variety of image prior models have been developed, such as transform based \cite{1,2,3}, total variation based \cite{4,5}, sparse representation based \cite{6,7} and nonlocal self-similarity based ones \cite{8,10,9,64}.

Transform based methods assume that natural images can be sparsely represented by some fixed basis (e.g., wavelet). Motivated by the fact, many wavelet shrinkage based methods have been proposed \cite{1,2,3}. For instance, Chang $\emph{et al}.$ \cite{1} proposed a method called Bayes shrink algorithm to model the wavelet transform coefficients as a generalized Gaussian distribution. Remenyi $\emph{et al}.$ \cite{3} attempted to use  2D scale mixing complex-value wavelet transform to improve denoising performance. In the total variation based methods \cite{4,5}, the image gradient is modeled as Laplacian distribution for image denoising.

Instead of modeling image statistics in some transform domain (e.g., gradient domain, wavelet domain), sparse representation based prior assumes that image patch can be precisely modeled as a sparse linear combination of basic elements. These elements, called atoms, compose a dictionary \cite{6,11,12,13}.  The seminal work of KSVD dictionary \cite{11} has not only confirmed promising denoising performance, but also been successfully used in various image processing and computer vision tasks \cite{14,15,16,65}. Nonetheless, there exist two main issues for typical patch-based sparse representation methods. First,  it is computationally expensive to learn an off-the-shelf dictionary; second, this kind of sparse representation model usually neglects the correlation between sparsely-coded patches.

Recently, a flurry of methods have exploited nonlocal self-similarity (NSS) prior based on the fact that natural images contain a large number of mutually similar patches at different locations. The seminal work of nonlocal means (NLM) \cite{8} utilized the NSS prior to implement a form of the weighted filtering for image denoising. Since then,  a flurry of nonlocal regularization methods were proposed to solve various image inverse problems \cite{17,18,19,20,21}. By contrast with the local regularization based methods (e.g., total variation method \cite{4}), nonlocal regularization based methods can effectively generate sharper image edges and preserve more image details. However, there are still lots of image details and structures that cannot be accurately recovered. One important reason is that the above nonlocal regularization terms rely on the weighted graph \cite{22}, and thus it is unavoidable that the weighted manner leads to disturbance and inaccuracy \cite{23}.

Inspired by the success of the NSS prior, recent studies \cite{9,10,24,25,26,27,71,72,40,59,56,63,64} have revealed that structured or group sparsity can provide more promising performance for noise removal. For instance, Dabov $\emph{et al}.$ \cite{10} proposed block matching and 3-D (BM3D) method to combine NSS prior and transform domain filtering,  which is still one of the state-of-the-art denoising methods. Marial $\emph{et al}.$ \cite{9} further advanced the idea of NSS by group sparse coding. Some other methods \cite{26,59,27,71,72} also have achieved highly competitive denoising results based on low rank property of the matrix formed by nonlocal similar patches in a natural image.

Though group sparsity has verified its great success in image denoising, most existing group-based sparse representation methods only consider the NSS prior of the noisy input. For example, LPG-PCA \cite{40} utilized nonlocal noisy similar patches as data samples to estimate statistical parameters for PCA training. NLGBT \cite{63} extracted the nonlocal similar patches from a noisy image and performed an iterative theresholding procedure to enforce group sparsity in the graph-based transform (GBT) domain. In SSC-GSM \cite{25}, the nonlocal similar patches are extracted from a noisy image by simultaneous sparse coding (SSC) and the group sparsity meets Gaussian scale mixture (GSM). However, such methods often suffer from a common drawback that the denoising performance may degrade quickly with increasing noise levels.

With the above question kept in mind, this paper proposes a new prior model for image denoising, called group sparse residual constraint (GSRC). Different from the previous group-based sparse representation denoising methods that only consider the single NSS prior of the noisy input, two kinds of NSS prior are (i.e.,  NSS prior of noisy and pre-filtered images) exploited for image denoising. The contribution of this paper is as follows. First, to enhance the performance of group-based sparse representation denoising methods, the group sparsity residual is proposed, and thus the problem of image denoising is transformed into one that reduces the group sparsity residual. Second, to reduce the residual, we first obtain some good estimation of the group sparse coefficients of the original image by pre-filtering and then the group sparse coefficients of the noisy  image are used to approximate this estimation. Third, we design an adaptive patch search scheme to improve the accuracy of the nonlocal similar patch selection. Fourth, to fuse these two NSS priors better, we present an effective iterative shrinkage algorithm to solve the proposed GSRC model. Experimental results show that the proposed GSRC modeling  outperforms many current state-of-the-art schemes such as BM3D \cite{10} and WNNM \cite{27}.

The reminder of this paper is organized as follows. Section~\ref{2} provides a  brief survey of the related work. Section~\ref{3} presents the modeling of group sparsity residual constraint (GSRC), adaptive patch search scheme, and discusses the main difference among the proposed GSRC method, the BM3D method \cite{10}, the NCSR method \cite{19} and  most  existing NSS prior-based denoising methods. Section~\ref{4} introduces the iterative shrinkage algorithm for solving the GSRC model. Section~\ref{5} presents the experimental results. Finally, Section~\ref{6} concludes this paper. The preliminary work has appeared in \cite{39}.

\section{Related Work}
\label{2}
Image denoising is a classical ill-posed inverse problem where the goal is to restore a latent clean image from its noisy observation. It has been widely recognized that the statistical modeling of natural image priors is crucial to the success of image denoising. Many image prior models have been developed in literature to characterize the statistical feature of natural images.

 Early models mainly consider the prior on level of pixels, such as the local structures used in Tikhonov regularization \cite{28} and total variation (TV) regularization \cite{4,5}. These methods are effective in removing the noise artifacts but smear out details and tend to over-smooth the images.

Another popular prior is based on image patch, which has shown promising performance in image denoising \cite{2,6,7,11}. The well-known work is sparse representation based model, which has been successfully exploited for image denoising \cite{6,7,11}. Sparse representation based model assumes that each patch of an image can be precisely represented by a sparse coefficient vector whose entries are mostly zero or close to zero based on a basis set called a dictionary. The dictionary is usually learned from a natural image dataset and the representative dictionary learning (DL) based methods (e.g., ODL \cite{12} and task driven DL \cite{13}) have been proposed and applied to image denoising and other image processing tasks.

Image patches that have similar patterns can be spatially far from each other and thus can be gathered in the whole image. The nonlocal self-similarity (NSS) prior characterizes the repetitiveness of textures and structures reflected by natural images within nonlocal regions, which can be exploited to retain the edges and the sharpness effectively. The seminal work of nonlocal means (NLM) denoising \cite{8} has motivated a wide range of studies on NSS and a flurry of NSS methods (e.g., BM3D \cite{10}, LSSC \cite{9} and NCSR \cite{19}) have been proposed and applied to image denoising tasks.

Low rank modeling based methods have been widely used and achieved great success in image or video denoising \cite{71,72,26,59,27}. A representative work was proposed by Ji $\emph{et al}.$ \cite{26}, to remove the flaws (e.g., noise, scratches and lines) in a video, the damaged pixels are first detected and demarcated as missing. The similar patches are grouped, satisfying that the patches in each group have similar underlying structure and carry out a low rank matrix approximately for each group. Finally, the matrix completion is conducted by each group to restore the image. Since the traditional low rank models tend to over-shrink the rank components and treat  different rank components equally, Gu $\emph{et al}.$ \cite{59,27} proposed the weighted nuclear norm minimization (WNNM) model for image denoising, which can achieve state-of-the-art  denoising performance.

Recently, deep learning based techniques for image denoising have been attracting considerable attentions due to its favorable denoising performance \cite{32,34,37,38,58,62}. For instance, Jain $\emph{et al}.$  \cite{32} proposed to use convolutional neural networks (CNNs) for image denoising and claimed that CNNs have similar or even better representation power than  Markov random field (MRF) model \cite{33}. In \cite{34}, the multi-layer perceptron (MLP) was successfully exploited for image denoising. Chen $\emph{et al}.$  \cite{62} proposed a trainable nonlinear reaction diffusion (TNRD) model for image denoising, which learned a modified fields of experts \cite{66} image prior by unfolding a fixed number of gradient descent inference steps. Zhang $\emph{et al}.$ \cite{37} investigated the construction of feed-forward denoising convolutional neural networks (Dn-CNN) to embrace the progress in very deep architecture, learning algorithm and regularization method into image denoising. Liu $\emph{et al}.$ \cite{38} considered the denoising problem as recursive image filtering via a hybrid neural network.
\section{Modeling of group sparsity residual constraint}
\label{3}
\subsection{Group-based Sparse Representation}

\begin{figure}[!htbp]
\begin{minipage}[b]{1\linewidth}
  \centering
  \centerline{\includegraphics[width=9cm]{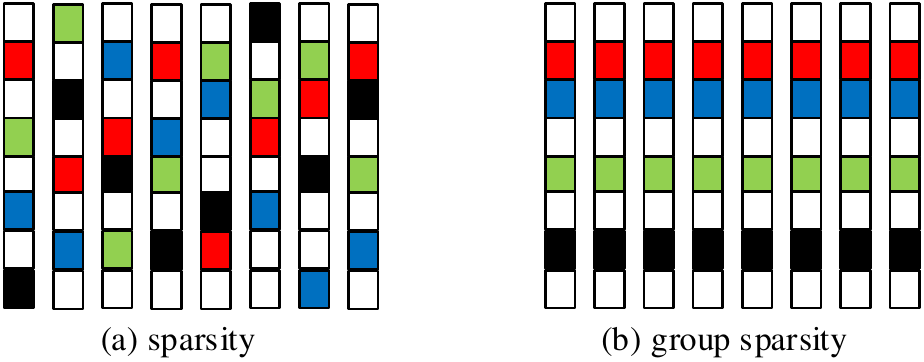}}
\end{minipage}
\caption{Comparison between sparsity (where columns are sparse, but do not alignment) and group sparsity (where  columns are sparse and aligned).}
\label{fig:1}
\end{figure}

Recent studies have revealed that structured or group sparsity can offer more promising performance for image restoration \cite{9,10,24,25,26,59,70,27,56,57,69}. Since the unit of our proposed sparse representation model is group, this subsection will give briefs to introduce how to construct the groups. To be concrete,  image  $\textbf{\emph{X}}$ with size $\emph{N}$ is divided into $\emph{n}$ overlapped patches $\textbf{\emph{x}}_i$ of size $\sqrt{b}\times\sqrt{b}, i=1,2,...,n$.  Then for each exemplar patch $\textbf{\emph{x}}_i$, its most similar $k$ patches are selected from an $L \times L$ sized searching window to form a set ${\textbf{\emph{S}}}_i$ (For the details of similar patch selection operator, please see subsection~\ref{3.3} ). After this, all the patches in ${\textbf{\emph{S}}}_i$ are stacked into a matrix ${\textbf{\emph{X}}}_i\in\Re^{{b}\times {k}}$, which contains every element of ${\textbf{\emph{S}}}_i$ as its column, i.e., ${\textbf{\emph{X}}}_i=\{{\textbf{\emph{x}}}_{i,1}, {\textbf{\emph{x}}}_{i,2}, ..., {\textbf{\emph{x}}}_{i,k}\}$.  The matrix ${\textbf{\emph{X}}}_i$ consisting of  all the patches with similar structures is called as a group, where ${\textbf{\emph{x}}_{i,k}}$ denotes the $k$-th similar patch (column form) of the $i$-th group. Finally, similar to patch-based sparse representation \cite{6,7,11}, given a dictionary ${\textbf{\emph{D}}}_i$, which is often learned from each group, such as DCT, PCA-based dictionary, each group ${\textbf{\emph{X}}}_i$ can be sparsely represented as ${\textbf{\emph{B}}}_i={{\textbf{\emph{D}}}_i}^{-1}\textbf{\emph{X}}_i$ and solved by the following $\ell_p$-norm minimization problem,
\begin{equation}
{{\textbf{\emph{B}}}_i}=\arg\min_{{\textbf{\emph{B}}}_i} \{||{\textbf{\emph{X}}}_i-{\textbf{\emph{D}}}_i{{\textbf{\emph{B}}}_i}||_F^2+\lambda_i||{{\textbf{\emph{B}}}_i}||_p\}
\label{eq:1}
\end{equation} 
where  $||\bullet||_F^2$ denotes the Frobenious norm,  $\lambda_i$ is the regularization parameter, and $p$  characterizes the sparsity of ${{\textbf{\emph{B}}}_i}$. Then the whole image ${\textbf{\emph{X}}}$ can be represented by the set of group sparse codes ${{\textbf{\emph{B}}}_i}$. Fig.~\ref{fig:1} shows the difference between sparsity and group sparsity.

In image denoising, the goal is to exploit group-based sparse representaion model to recover ${\textbf{\emph{X}}}_i$ from noisy observation ${\textbf{\emph{Y}}}_i$ and solve the following minimization problem,
\begin{equation}
{{\textbf{\emph{A}}}_i}=\arg\min_{{\textbf{\emph{A}}}_i} \{||{\textbf{\emph{Y}}}_i-{\textbf{\emph{D}}}_i{{\textbf{\emph{A}}}_i}||_F^2+\lambda_i||{{\textbf{\emph{A}}}_i}||_p\}
\label{eq:2}
\end{equation} 

Once all group sparse codes ${{\textbf{\emph{A}}}_i}$ are achieved, the latent clean image can be reconstructed as $\hat{\textbf{\emph{X}}}={\textbf{\emph{D}}}{\textbf{\emph{A}}}$, where ${\textbf{\emph{A}}}$ includes the set of group sparse codes ${{\textbf{\emph{A}}}_i}$.
Although group sparsity has demonstrated its effectiveness in image denoising, most existing group-based sparse representation denoising methods only use the NSS prior of noisy  image for noise removal (e.g., Eq.~\eqref{eq:2}), What's more, the denoising performance may degrade quickly with increasing noise levels, making it challenging to recover the latent clean image directly from its noisy observation.
 \begin{figure*}[!htbp]
\begin{minipage}[b]{1\linewidth}
  \centering
  \centerline{\includegraphics[width=18cm,height=7cm]{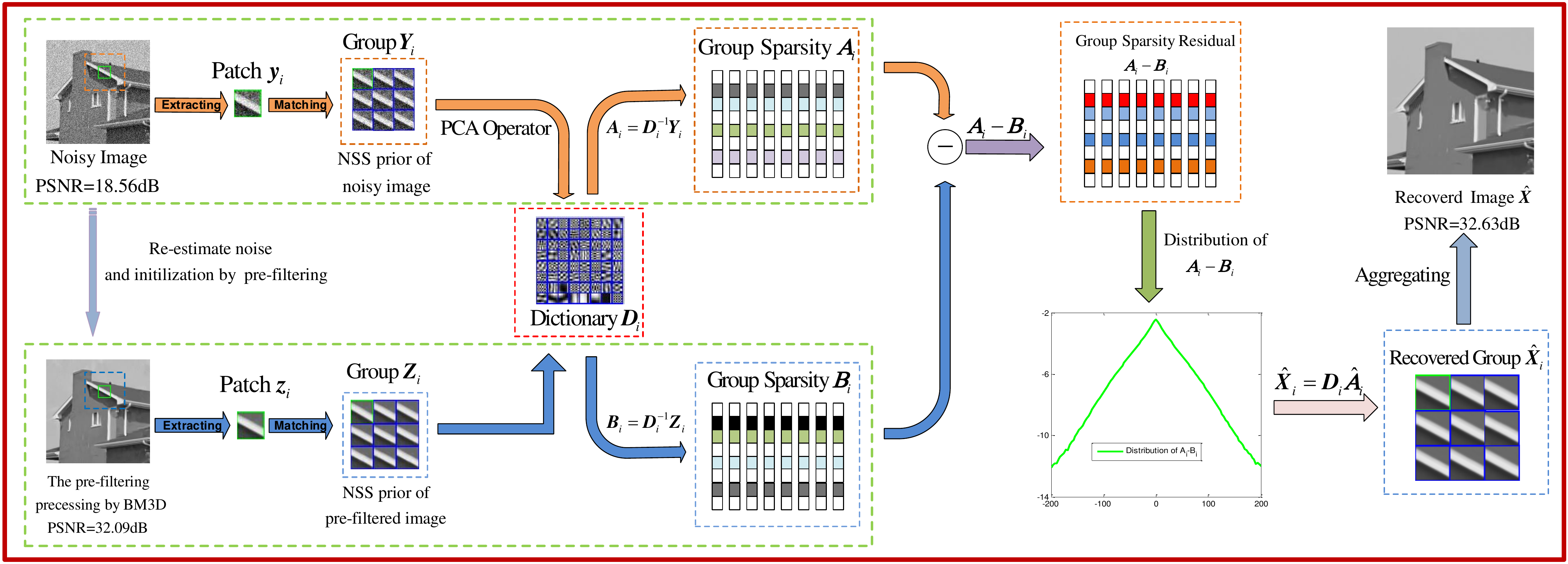}}
\end{minipage}
\caption{Flowchart of image denoising by  group sparsity residual constraint (GSRC) model.}
\label{fig:2}
\end{figure*}
\subsection{Group Sparsity Residual Constraint}
Let us revisit Eq.~\eqref{eq:1} and Eq.~\eqref{eq:2}, due to the influence of noise, it is very difficult to estimate the true group sparse code ${{\textbf{\emph{B}}}}$ from noisy image ${{\textbf{\emph{Y}}}}$. In other words, the group sparse code ${{\textbf{\emph{A}}}}$ obtained by solving Eq.~\eqref{eq:2} is expected to be close enough to the true group sparse code ${{\textbf{\emph{B}}}}$ of the original image ${{\textbf{\emph{X}}}}$ in Eq.~\eqref{eq:1}. As a consequence, the quality of image denoising largely depends on the level of the group sparsity residual,  which is defined as the difference between group sparse code ${{\textbf{\emph{A}}}}$ and true group sparse code ${{\textbf{\emph{B}}}}$,
\begin{equation}
{{\textbf{\emph{R}}}}={{\textbf{\emph{A}}}}-{{\textbf{\emph{B}}}}
\label{eq:3}
\end{equation} 
Therefore, to reduce the group sparsity residual ${{\textbf{\emph{R}}}}$ and boost the accuracy of  ${{\textbf{\emph{A}}}}$, we propose a new prior model to image denoising, called group sparse residual constraint (GSRC) \cite{39}, and  thus Eq.~\eqref{eq:2} can be rewritten as
\begin{equation}
{{\textbf{\emph{A}}}_i}=\arg\min_{{\textbf{\emph{A}}}_i} \{||{\textbf{\emph{Y}}}_i-{\textbf{\emph{D}}}_i{{\textbf{\emph{A}}}_i}||_F^2+\lambda_i||{{\textbf{\emph{A}}}_i}-{{\textbf{\emph{B}}}_i}||_p\}
\label{eq:4}
\end{equation} 

However, it can be seen that the true group sparse code ${{\textbf{\emph{B}}}}$ and $p$ are unknown since the original image ${{\textbf{\emph{X}}}}$ is not available. Therefore, we will discuss how to obtain ${{\textbf{\emph{B}}}}$ and $p$. In addition, one important issue of the proposed GSRC based image denoising is the selection of the dictionary. To adapt to the local image structures, instead of learning an over-complete dictionary for each group ${{\textbf{\emph{Y}}}}_i$ as in \cite{9}, we learn the principle component analysis (PCA) based dictionary \cite{19} for each group ${{\textbf{\emph{Y}}}}_i$.
\subsection{Estimation of the Unknown Group Sparse Code}
Eq.~\eqref{eq:3} shows that by reducing the group sparsity residual $\textbf{\emph{R}}$,  we could improve the performance of image denoising. In general, the original image $\textbf{\emph{X}}$ is not available in practice, and thus the true group sparse code $\textbf{\emph{B}}$ is unknown. However,  the true group sparse code $\textbf{\emph{B}}$ can be estimated based on prior knowledge of the original image ${{\textbf{\emph{X}}}}$ we have. For example, if we have many example images  similar to the original image ${{\textbf{\emph{X}}}}$, then  a good estimation of ${{\textbf{\emph{B}}}}$ could be learned from the example image set. However, under many practical situations, the example image set is simply and unsuitable.

The strategy of pre-filtering is a popular means to image denoising. The basic idea is similar to many denoising algorithms such as BM3D \cite{10} where a first stage pilot denoising is exploited before going to the second stage of the actual denoising. In past few years, a variety of image denoising methods based on pre-filtering have been developed, such as LPG-PCA \cite{40}, TID \cite{41}, SOS \cite{42}, and aGMM \cite{43} methods, etc.

Based on the above analysis, we first apply pre-filtering (e.g., BM3D \cite{10}, EPLL \cite{44}) to noisy image $\textbf{\emph{Y}}$, and then the initialization result of pre-filtering is defined as $\textbf{\emph{Z}}$. Since the pre-filtering has an ideal denoising performance, $\textbf{\emph{Z}}$ could be regarded as a good approximation of the original image $\textbf{\emph{X}}$. Therefore, in this paper the group sparse code $\textbf{\emph{B}}$ is achieved by the pre-filtering $\textbf{\emph{Z}}$. The flowchart of the proposed GSRC is illustrated in Fig.~\ref{fig:2}. Specifically, to reduce the group sparsity residual, we first obtain a good estimation of the group sparse coefficients of the original image by pre-filtering $\textbf{\emph{Z}}$ and then the group sparse coefficients of noisy input image  are used to approximate this estimate.
\subsection{Adaptive Patch Search Scheme}
\label{3.3}
\begin{figure}[!htbp]
\begin{minipage}[b]{1\linewidth}
\centering
\centerline{\includegraphics[width=9cm]{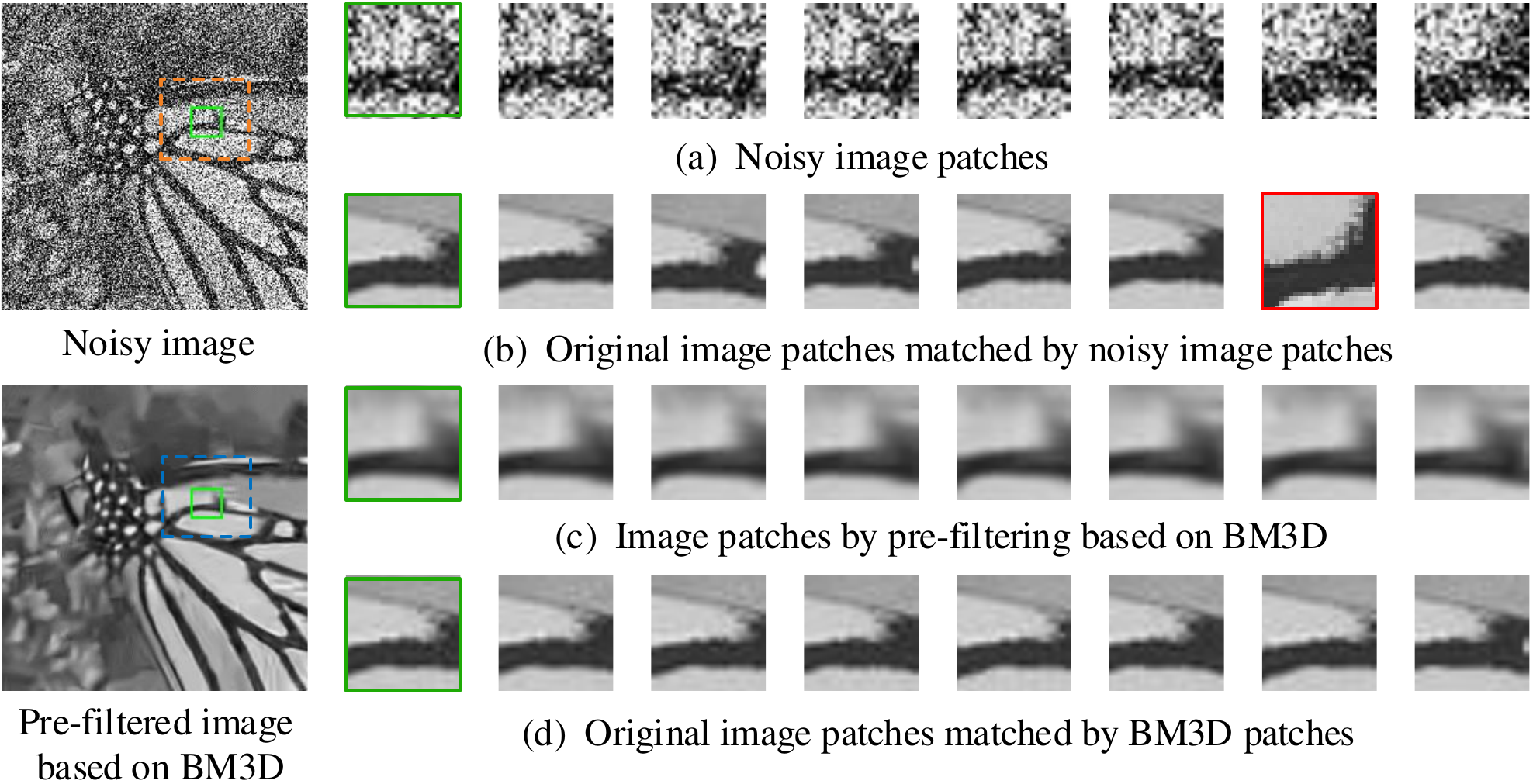}}
\end{minipage}
\caption{Patch selection between noisy image and pre-filtered image based on BM3D via $k$NN method (where green box represents the reference patch).}
\label{fig:3}
\end{figure}

$k$ Nearest Neighbors ($k$NN) method \cite{45} has been widely used to nonlocal similar patch selection. Given a noisy reference patch and a target dataset, the aim of $k$NN is to find the $k$ most similar patches. However, since the given reference patch is noisy, $k$NN has a drawback that some of the $k$ selected patches may not be truly similar to given reference patch. For instance, the noisy similar patches via $k$NN and the clean patches matched with these noisy similar patch indexes are shown in Fig.~\ref{fig:3}(a) and Fig.~\ref{fig:3}(b), respectively. It can be seen  that the 7-th patch (red box) is obviously deviating from given reference patch (green box) in Fig.~\ref{fig:3}(b). Since the pre-filtered image is regarded as a good estimation of the original image, in this paper we first adopt  pre-filtering result as the target image to fetch the $k$ most similar patch indexes. Fig.~\ref{fig:3}(c) shows the similar patches  of BM3D-based pre-filtered image  searched by $k$NN and Fig.~\ref{fig:3}(d) shows the clean patches matched with the pre-filtered image similar patch indexes. It can be seen that the similar patch selection of the pre-filtered image is  more accurate  than that of the noisy image. Therefore, to obtain an effective similar patch indexes via $k$NN, an adaptive patch search scheme is designed. We define the following formula,
\begin{equation}
\partial ={\rm SSIM} ({{\textbf{\emph{Z}}}}, {\hat{{\textbf{\emph{X}}}}}^{\ell+1})-{\rm SSIM} ({{\textbf{\emph{Z}}}}, {\hat{{\textbf{\emph{X}}}}}^{\ell})
\label{eq:5}
\end{equation} 
where SSIM represents structural similarity \cite{46} and ${\hat{{\textbf{\emph{X}}}}}^{\ell}$ represents the $\ell$-th iteration denoising result. We empirically define that if $\partial<\tau$,  ${\hat{{\textbf{\emph{X}}}}}^{\ell+1}$ is regarded as target image to fetch the $k$ similar patch indexes, otherwise  ${{\textbf{\emph{Z}}}}$ is regarded as target image. ${{\textbf{\emph{Z}}}}$ is the pre-filtered image and $\tau$ is a small constant.
\subsection{Discussion}
This subsection will provide  detailed discussion about the main difference among  the proposed GSRC method, the BM3D mehod\cite{10}, the NCSR method \cite{19} and  most  existing NSS prior-based denoising methods.
\begin{itemize}
\item {It can be seen that the proposed GSRC method is similar to BM3D method, both of them are two stage-based denoising methods. Nonetheless, the BM3D is based on filtering method (e.g., DST, DCT), while the proposed GSRC is based on sparse coding method, along with dictionary learning. Compared with the analytically designed dictionaries (e.g., DCT/wavelet dictionary), dictionaries learned from image patch/group have an advantage of being better adapted to image local structures \cite{6,11}, and thus could enhance the sparsity which leads to better performance. Moreover, in the second stage of BM3D, they are mixing up the pre-filtered image group with the noisy image group unmannerly, and the Winner filtering is utilized. However, we propose a new prior model, called group sparsity residual constraint (GSRC). Unlike the BM3D method, we do not mix up the pre-filtering image group with the noisy image group, instead, we adopt an iterative shrinkage algorithm \cite{49} to solve the proposed GSRC model, which can better integrate these two NSS priors of noisy and pre-filtered images (Section~\ref{4} for more details).}

\item {Natural images  often possess similar repetitive patterns, i.e., a large number of nonlocal redundancies \cite{8}. By searching many nonlocal patches similar to given reference patch, NCSR \cite{19} first obtained  good estimates of the sparse coding coefficients of the original image by the principle of NLM, and then centralized the sparse coding coefficients of the observed image to those estimates to improve the performance of denoising. However, due to the fact that NLM depends on the weighted graph \cite{22}, it is unavoidable that the weighted manner leads to disturbance and inaccuracy \cite{23}. It is worth mentioning that the proposed GSRC model does not involve in the weighted graph. In addition,  NCSR is actually a patch-based sparse representation method, which usually neglects the relationships among  similar patches \cite{24,72}.}

\item NSS prior has shown great success in image denoising. Most existing denoising methods only exploit the NSS prior of noisy  image \cite{9,19,25,26,59,27,47}, and few  methods use the NSS prior from natural images \cite{48}. Actually, different from  the most existing NSS prior-based denoising methods, in this work  we  consider two kinds of NSS prior, i.e., NSS priors of noisy and pre-filtered images. Experimental results show that the proposed GSRC scheme  outperforms many state-of-the-art methods, such as BM3D \cite{10} and WNNM \cite{27} (See Section~\ref{5} for more details).
\end{itemize}
\begin{figure}[!htbp]
\begin{minipage}[b]{1\linewidth}
\centering
\centerline{\includegraphics[width=9cm]{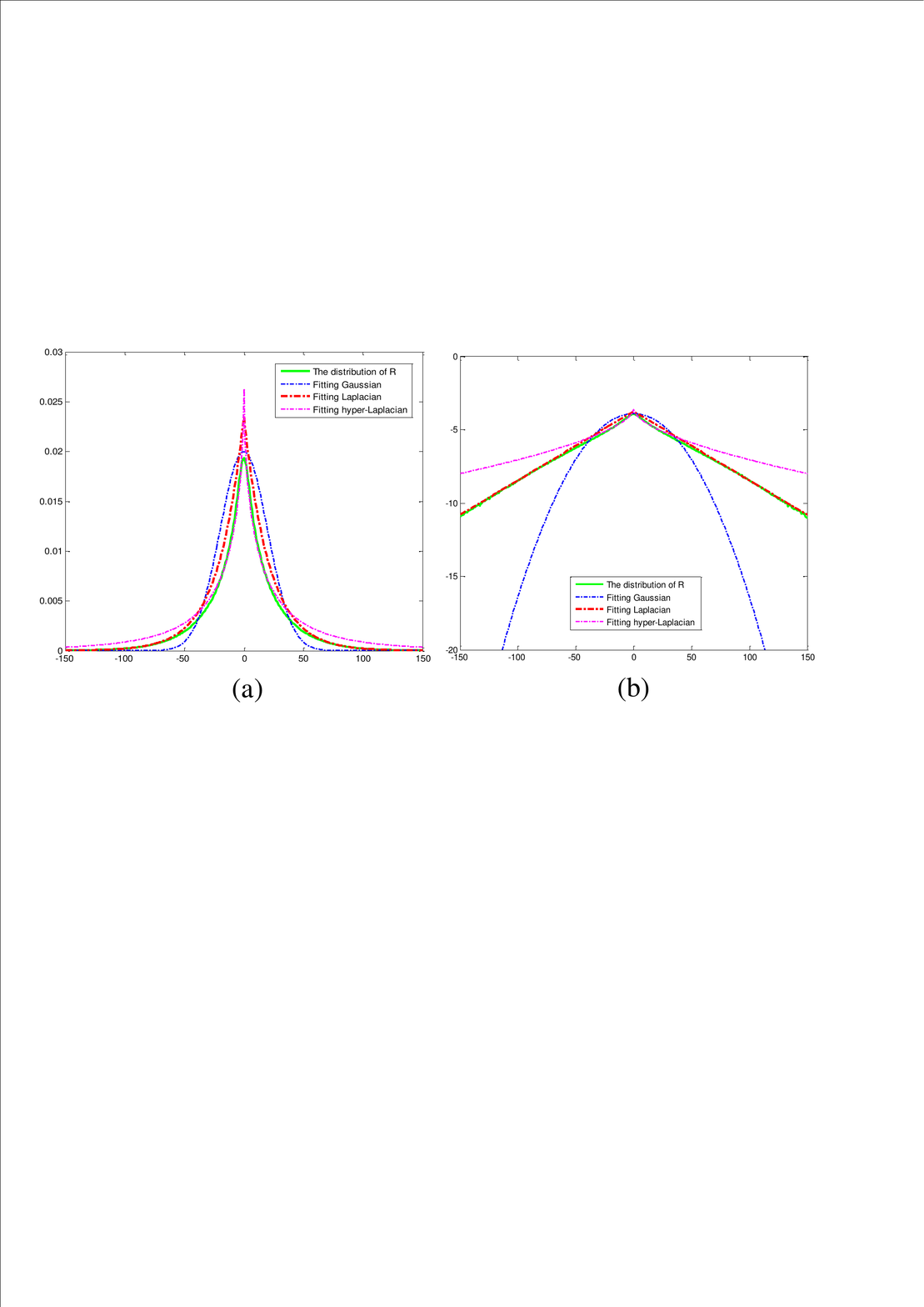}}
\end{minipage}
\caption{The distribution of the group sparsity residual ${{\textbf{\emph{R}}}}$ for image $\emph{lena}$ with $\sigma$=30 and fitting Gaussian, Laplacian and hyper-Laplacian distribution in (a) linear and (b) log domain, respectively (pre-filtering based on BM3D \cite{10}).}
\label{fig:4}
\end{figure}
\begin{figure}[!htbp]
\begin{minipage}[b]{1\linewidth}
\centering
\centerline{\includegraphics[width=9cm]{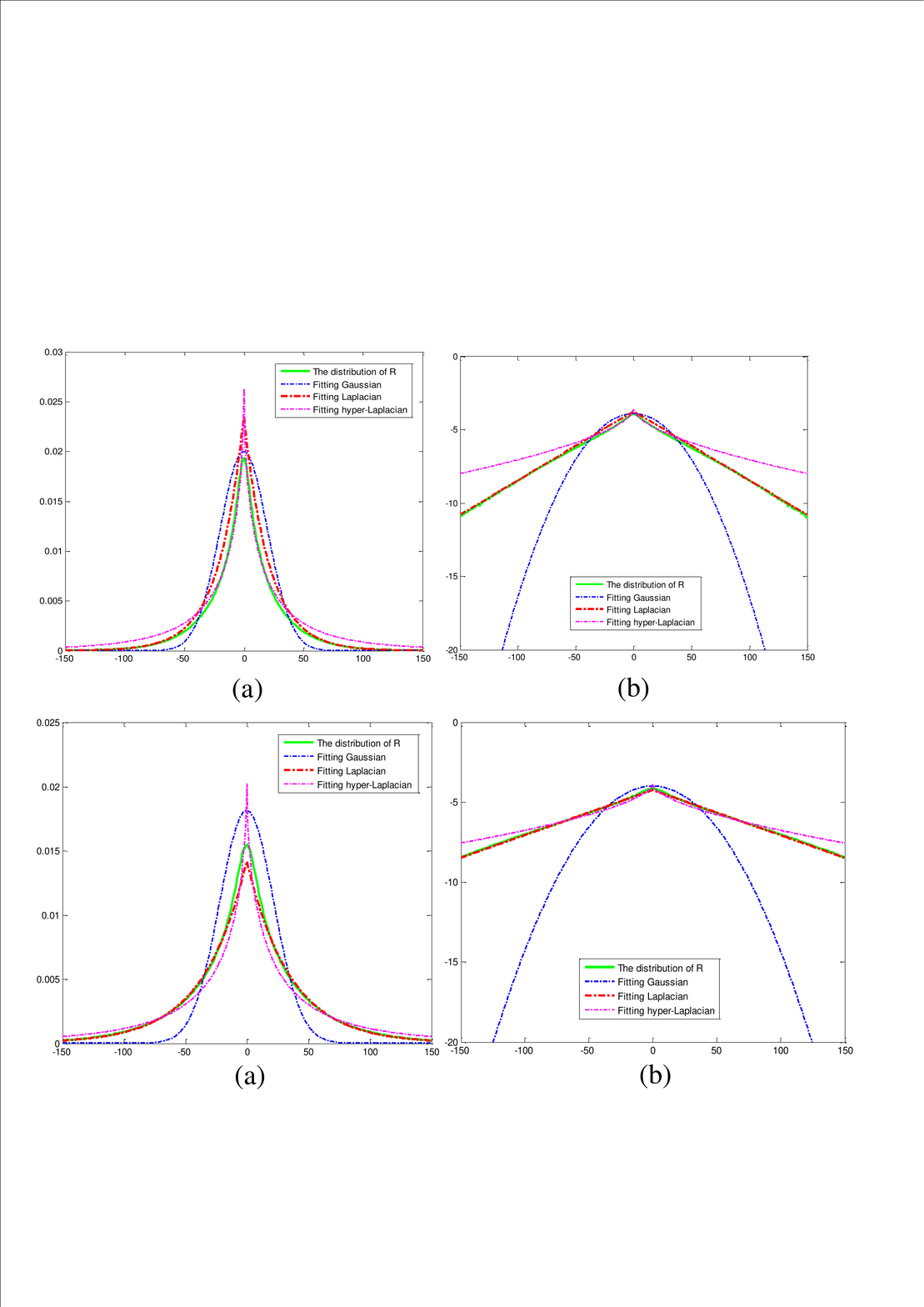}}
\end{minipage}
\caption{The distribution of the group sparsity residual ${{\textbf{\emph{R}}}}$ for image $\emph{House}$ with $\sigma$=50 and fitting Gaussian, Laplacian and hyper-Laplacian distribution in (a) linear and (b) log domain, respectively (pre-filtering based on EPLL \cite{44}).}
\label{fig:5}
\end{figure}
\begin{figure*}[!htbp]
\begin{minipage}[b]{1\linewidth}
\centering
\centerline{\includegraphics[width=18cm]{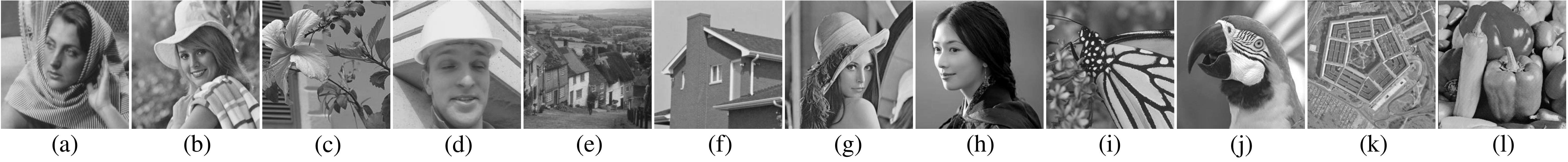}}
\end{minipage}
\caption{The 12 test images for denoising experiments. (a) Barbara; (b) Elaine; (c) flower; (d) foreman; (e) Hill; (f) House; (g) lena; (h) lin; (i) Monarch; (j) Parrot; (k) pentagon; (l) peppers.}
\label{fig:6}
\end{figure*}
\section{Algorithm of GSRC}
\label{4}
\subsection{Setting of the Parameter $p$}
In Eq.~\eqref{eq:4}, except for estimating ${{\textbf{\emph{B}}}}$, we also need to set the value of $p$. Here we perform some experiments to investigate the statistical property of the group sparsity residual ${{\textbf{\emph{R}}}}$, where ${{\textbf{\emph{R}}}}$ represents the set of ${{\textbf{\emph{R}}}}_i={{\textbf{\emph{A}}}}_i-{{\textbf{\emph{B}}}}_i$. In these experiments, two images $\emph{lena}$ and $\emph{House}$ are used as examples, where Gaussian white noise is added to the images $\emph{lena}$ and $\emph{House}$  with standard deviation $\sigma$= 30 (pre-filtering based on BM3D) and $\sigma$= 50 (pre-filtering based on EPLL), respectively. We plot the histogram of ${{\textbf{\emph{R}}}}$ as well as the fitting Gaussian, Laplacian and hyper-Laplacian distribution of ${{\textbf{\emph{R}}}}$ in Fig.~\ref{fig:4}(a) and Fig.~\ref{fig:5}(a). To better observe the fitting of the tails, we also plot these distributions in the log domain in Fig.~\ref{fig:4}(b) and Fig.~\ref{fig:5}(b). It can be seen that the histogram of ${{\textbf{\emph{R}}}}$ can be well characterized  by the Laplacian distribution. Therefore, we set $p=1$ and the $\ell_1$-norm is adopted to regularize each group sparsity residual ${{\textbf{\emph{R}}}_i}$, and Eq.~\eqref{eq:4} can be rewritten as
\begin{equation}
\begin{aligned}
{{\textbf{\emph{A}}}_i}& = \arg\min_{{\textbf{\emph{A}}}_i} \{||{\textbf{\emph{Y}}}_i-{\textbf{\emph{D}}}_i{{\textbf{\emph{A}}}_i}||_F^2+\lambda_i||{{\textbf{\emph{A}}}_i}-{{\textbf{\emph{B}}}_i}||_1\}\\
& =\arg\min_{\tilde{\boldsymbol\alpha}_i} \{||\tilde{{\textbf{\emph{y}}}}_i-\tilde{{\textbf{\emph{D}}}}_i{\tilde{\boldsymbol\alpha}_i}||_2^2+\lambda_i||{\tilde{\boldsymbol\alpha}_i}-{\tilde{\boldsymbol\beta}_i}||_1\}
\end{aligned}
\label{eq:6}
\end{equation} 
where $\tilde{{\textbf{\emph{y}}}}_i, {\tilde{\boldsymbol\alpha}_i}$, and ${\tilde{\boldsymbol\beta}_i}$ denote the vectorization of the matrix ${{\textbf{\emph{Y}}}_i}, {{\textbf{\emph{A}}}_i}$ and ${{\textbf{\emph{B}}}_i}$, respectively. Each column $\tilde{{\textbf{\emph{d}}}}_j$ of the matrix $\tilde{{\textbf{\emph{D}}}}_i=[\tilde{{\textbf{\emph{d}}}}_1, \tilde{{\textbf{\emph{d}}}}_2, ..., \tilde{{\textbf{\emph{d}}}}_J]$  denotes the vectorization of the rank-one matrix, where $J$ denotes the number of dictionary atoms.

\subsection{Iterative Shrinkage Algorithm to Solve the Proposed GSRC Model}
For fixed ${\tilde{\boldsymbol\beta}_i}, \lambda_i$, Eq.~\eqref{eq:6} is convex and can be solved efficiently. We adopt an iterative shrinkage algorithm in \cite{49} to solve Eq.~\eqref{eq:6}. In the $\ell+1$-iteration, the proposed shrinkage operator can be calculated as
\begin{equation}
{\tilde{\boldsymbol\alpha}_i}^{\ell+1}= {{{\textbf{\emph{S}}}}_{\lambda_i}}({\tilde{{\textbf{\emph{D}}}}_i}^{-1}{{\hat{\tilde{{\textbf{\emph{x}}}}}}_i}^{\ell}-{{\tilde{\boldsymbol\beta}_i}})+{{\tilde{\boldsymbol\beta}_i}}
\label{eq:7}
\end{equation} 
where ${{{\textbf{\emph{S}}}}_{\lambda_i}}(\cdot)$ is the soft-thresholding operator, ${{\hat{\tilde{{\textbf{\emph{x}}}}}}_i}$  represents the vectorization of the $i$-th reconstructed group ${{\hat{{\textbf{\emph{X}}}}}_i}$. In fact, according to Eq.~\eqref{eq:7}, one can observe that these two NSS priors can be better integrated into this surrogate algorithm.  The above shrinkage operator follows the standard surrogate algorithm, from which more details can be seen in \cite{49}.

\subsection{Adaptive Group Sparsity Regularization Parameter Setting}
The parameter $\lambda_i$ for each group that balances the fidelity term and the regularization term should be adaptively determined for better denoising performance. In this subsection, inspired by \cite{1}, we propose a more robust method for computing $\lambda_i$ of Eq.~\eqref{eq:6} by formulating the group sparse estimation as a Maximum A-Posterior (MAP) estimation problem. For a given ${{\textbf{\emph{B}}}_i}$, the optimal solution of Eq.~\eqref{eq:6} is ${\hat{{{\textbf{\emph{R}}}}}}_i=\arg\max\limits_{{\textbf{\emph{R}}}_{\emph{i}}}\rm log\ \emph{P}({{\textbf{\emph{R}}}_{\emph{i}}}|{\textbf{\emph{Y}}}_{\emph{i}})$. By Bayes' formula, it is equivalent to
\begin{equation}
\begin{aligned}
{\hat{{{\textbf{\emph{R}}}}}}_i & =\arg\max_{{\textbf{\emph{R}}}_{\emph{i}}}\{\rm log\ \emph{P}({{\textbf{\emph{R}}}_{\emph{i}}}|{\textbf{\emph{Y}}}_{\emph{i}})\}\\
&=\arg\min\limits_{{\textbf{\emph{R}}}_{\emph{i}}}\{-\rm log\ \emph{P}({\textbf{\emph{Y}}}_{\emph{i}} |{{\textbf{\emph{R}}}_{\emph{i}}})-\rm log\ \emph{P}({{\textbf{\emph{R}}}_{\emph{i}}})\}
\end{aligned}
\label{eq:8}
\end{equation}

The log-likelihood term $\rm log\ \emph{P}({{\textbf{\emph{R}}}_{\emph{i}}}|{\textbf{\emph{Y}}}_{\emph{i}})$ is characterized by the statistics of noise $\textbf{\emph{V}}$, which is assumed to be additive white Gaussian noise with standard deviation $\sigma$, and thus we have
\begin{equation}
\begin{aligned}
\emph{P}({\textbf{\emph{Y}}}_{\emph{i}} |{{\textbf{\emph{R}}}_{\emph{i}}})=\emph{P}({\textbf{\emph{Y}}}_{\emph{i}} |{{\textbf{\emph{A}}}_{\emph{i}}}, {{\textbf{\emph{B}}}_{\emph{i}}}) & = {\rm exp}(-\frac{1}{2{\sigma}^2}||{\textbf{\emph{Y}}}_{\emph{i}}-{{\textbf{\emph{D}}}_{\emph{i}}}{{\textbf{\emph{A}}}_{\emph{i}}}||_F^2)
\end{aligned}
\label{eq:9}
\end{equation}
where ${{\textbf{\emph{R}}}_{\emph{i}}}$ and ${{\textbf{\emph{B}}}_{\emph{i}}}$ are assumed to be independent. Since the group sparsity residual ${{\textbf{\emph{R}}}_{\emph{i}}}$ can be well characterized by the Laplacian distribution from Fig.~\ref{fig:4} and Fig.~\ref{fig:5}. Thus, the prior distribution $\emph{P}({{\textbf{\emph{R}}}_{\emph{i}}})$ is characterized by an i.i.d Laplacian distribution,
\begin{equation}
\emph{P}({{\textbf{\emph{R}}}_{\emph{i}}})=\frac{c}{\sqrt{2}\sigma_{\emph{i}}}\rm exp(-\frac{\emph{c}\sqrt{2}|{{\textbf{\emph{R}}}_{\emph{i}}}|}{\sigma_{\emph{i}}})
\label{eq:10}
\end{equation}

Then we substitute Eq.~\eqref{eq:9} and Eq.~\eqref{eq:10} into Eq.~\eqref{eq:8}, and thus we can readily derive the desired regularization parameter $\lambda_i$ for each group,
 \begin{equation}
\lambda_i=\frac{c*2\sqrt{2}{\sigma}^2}{\sigma_i}
\label{eq:11}
\end{equation}
where $\sigma_i$ denotes the estimated variance of each group sparsity residual ${\textbf{\emph{R}}}_i$, and $c$ is a small constant.

\begin{table}[!htbp]
\caption{Group sparsity residual constraint for image denoising.}
\centering  
\begin{tabular}{lccc}  
\hline  
$\textbf{Input:}$ \ Noisy image ${{\textbf{\emph{Y}}}}$.\\
  $\rm \textbf{Initialization:} \  {\hat{{\textbf{\emph{X}}}}}={{\textbf{\emph{Y}}}}, {{\textbf{\emph{Z}}}}, \emph{c}, \emph{k}, \emph{b}, \emph{L}, \sigma, \tau, \gamma, \delta$;\\
  $\rm \textbf{For}$\ $\ell=1, 2, ..., K$ $\rm \textbf{do}$\\
  \qquad Iterative regularization ${{\textbf{\emph{Y}}}}^{\ell+1}= {\hat{\textbf{\emph{X}}}}^{\ell}+\delta({{\textbf{\emph{Y}}}}-{\hat{\textbf{\emph{X}}}}^{\ell})$;\\
    \qquad Re-estimate  $\sigma^{\ell+1}$ computing by Eq.~\eqref{eq:12};\\
\qquad $\textbf{If}$ \ \ $\ell=1$\\
       \qquad \qquad Similar patch indexes selection  based on ${{\textbf{\emph{Z}}}}$.\\
\qquad $\textbf{Else}$\\
  \qquad\qquad $\textbf{If}$ \ \ ${\rm SSIM}({{\textbf{\emph{Y}}}}^{\ell+1}, {{\textbf{\emph{Z}}}})-{\rm SSIM}({{\textbf{\emph{Y}}}}^{\ell}, {{\textbf{\emph{Z}}}})<\tau$\\
  \qquad \qquad  \qquad Similar patch indexes selection  based on ${{\textbf{\emph{Y}}}}^{\ell+1}$.\\
  \qquad \qquad  $\textbf{Else}$\\
   \qquad \qquad \qquad Similar patch indexes selection  based on ${{\textbf{\emph{Z}}}}$.\\
   \qquad  \qquad $\textbf{End if}$\\
\qquad $\textbf{End if}$\\
 \qquad $\rm \textbf{For}$\ each patch ${{\textbf{\emph{y}}}}_i$ and ${{\textbf{\emph{z}}}}_i$ $\rm \textbf{do}$\\
   \qquad \qquad  Find a group ${{{\textbf{\emph{Y}}}}_i}^{\ell+1}$ via $k$NN.\\
   \qquad \qquad Find a group ${{{\textbf{\emph{Z}}}}_i}^{\ell+1}$ via $k$NN.\\
   \qquad \qquad Constructing dictionary ${{{\textbf{\emph{D}}}}_i}^{\ell+1}$ by ${{{\textbf{\emph{Y}}}}_i}^{\ell+1}$ by PCA operator.\\
   \qquad \qquad Update ${{{\textbf{\emph{B}}}}_i}^{\ell+1}$ computing by ${{\textbf{\emph{B}}}}_i={{{\textbf{\emph{D}}}}_i}^{-1}{{\textbf{\emph{Z}}}}_i$.\\
   \qquad \qquad Update ${\lambda_i}^{t+1}$ computing by Eq.~\eqref{eq:11}.\\
   \qquad \qquad Update ${{{\textbf{\emph{A}}}}_i}^{\ell+1}$ computing by Eq.~\eqref{eq:7}.\\
   \qquad \qquad Get the estimation ${{{\textbf{\emph{X}}}}_i}^{\ell+1}$ =${{{\textbf{\emph{D}}}}_i}^{\ell+1}$${{{\textbf{\emph{A}}}}_i}^{\ell+1}$.\\
  \qquad  $\rm \textbf{End for}$\\
   \qquad \qquad Aggregate ${{{\textbf{\emph{X}}}}_i}^{\ell+1}$ to form the recovered image ${\hat{{\textbf{\emph{X}}}}}^{\ell+1}$.\\
    $\rm \textbf{End for}$\\
     $\textbf{Output:}$ ${\hat{\textbf{\emph{X}}}}^{\ell+1}$.\\
\hline
\end{tabular}
\label{lab:1}
\end{table}

With the solution ${{\textbf{\emph{A}}}_i}$ in Eq.~\eqref{eq:7}, the clean group ${{\textbf{\emph{X}}}_i}$ can be reconstructed as ${{\hat{\textbf{\emph{X}}}}_i}={{\textbf{\emph{D}}}_i}{{\textbf{\emph{A}}}_i}$. Then the latent clean image ${{\hat{\textbf{\emph{X}}}}}$ can be reconstructed by aggregating all the groups $\{{{\textbf{\emph{X}}}_i}\}$. In practical, we could perform the above denoising procedures for better results by several iterations. In the $\ell$-th iteration, the iterative regularization strategy \cite{50} is used to update the estimation of noise variance. Then the standard deviation of noise in $\ell$-th iteration is adjusted as
 \begin{equation}
{\sigma^{\ell}}=\gamma*\sqrt{({\sigma^2-||{{\textbf{\emph{Y}}}}-{\hat{{\textbf{\emph{X}}}}}^{\ell}||_2^2})}
\label{eq:12}
\end{equation}
where $\gamma$ is a constant. The complete description of the proposed method for image denoising based on GSRC model is exhibited in Table~\ref{lab:1}.

\section{Experimental Results}
\label{5}
In this section, extensive experimental results are presented to evaluate the denoising performance of the proposed GSRC. For the test images, we use two different test datasets for thorough evaluation. One is a test dataset containing 200 natural images from Berkeley segmentation dataset (BSD200) \cite{51} and the other one contains 12 images which are shown in Fig.~\ref{fig:6}. We consider two versions of pre-filtering: (1) a pre-filtered image $\textbf{\emph{Z}}$ generated by the BM3D method \cite{10}, denoted as GSRC-BM3D;  (2) a pre-filtered image $\textbf{\emph{Z}}$ generated by the EPLL method \cite{44}, denoted as GSRC-EPLL. To evaluate the quality of denoised image, both PSNR and SSIM \cite{46} metrics are used. 

\subsection{Parameter Setting}
Parameters used in the algorithm are empirically chosen in consideration of the noise levels in order to achieve relatively good performance. The basic parameter setting is as follows: the searching window $L \times L$ is set to be $30 \times 30$. The size of patch $\sqrt{b} \times \sqrt{b}$ is set to be $6 \times6$, $7 \times7$, $8 \times8$ and $9 \times9$ for $\sigma \leq 20$, $20<\sigma \leq 50$, $50<\sigma \leq 75$ and $75<\sigma \leq 100$, respectively. The searching matched patches $k$ is set to be 60, 80, 90 for $\sigma \leq 50$, $50<\sigma \leq 75$ and $75<\sigma \leq 100$, respectively. The detailed setting of the involved parameters $c$, $\delta$, $\gamma$ and $\tau$ are shown in Table~\ref{lab:2.0}. We run denoising experiments for a large range of noise standard deviations ($\sigma$= 20, 30, 40, 50, 75 and 100).
\begin{table}[!htbp]
\caption{The detailed involved parameters setting of $c, \delta, \gamma, \tau$.}
\centering  
\begin{tabular}{|c|c|c|c|c||c|c|c|c|}
\hline\hline
  \multicolumn{1}{|c|}{Noise level}&\multicolumn{4}{|c||}{GSRC-BM3D}&\multicolumn{4}{|c|}{GSRC-EPLL}\\
\hline
\multirow{1}{*}{{{$\sigma$}}}&\multirow{1}{*}{{{c}}}&\multirow{1}{*}{{{$\delta$}}}&\multirow{1}{*}{\textbf{{$\gamma$}}}
&{{{$\tau$}}}&\multirow{1}{*}{{{c}}}&\multirow{1}{*}{{{$\delta$}}}&\multirow{1}{*}{\textbf{{$\gamma$}}}
&{{{$\tau$}}}\\
\hline
 \multirow{1}{*}{$\sigma\leq20$}     &    0.2    &  0.2      & 0.7      & 1e-4      & 0.3      & 0.1     & 0.5     & 5e-4\\
\hline
 \multirow{1}{*}{$20<\sigma\leq30$}  &    0.4    &  0.1      & 0.5      & 7e-4      & 0.3      & 0.1     & 0.5     & 5e-4\\
\hline
 \multirow{1}{*}{$30<\sigma\leq40$}  &    0.2    &  0.2      & 0.7      & 6e-5      & 0.3      & 0.1     & 0.5     & 6e-4\\
\hline
 \multirow{1}{*}{$40<\sigma\leq50$}  &    0.5    &  0.1      & 0.4      & 6e-5      & 0.5      & 0.1     & 0.4     & 4e-4\\
\hline
 \multirow{1}{*}{$50<\sigma\leq75$}  &    0.9    &  0.1      & 0.3      & 6e-5      & 0.9      & 0.1     & 0.3     & 1e-4\\
\hline
 \multirow{1}{*}{$75<\sigma\leq100$} &    1      &  0.1      & 0.3      & 2e-4      & 0.9      & 0.1     & 0.3     & 2e-4\\
\hline\hline
\end{tabular}
\label{lab:2.0}
\end{table}
\begin{table*}[!htbp]
\caption{PSNR ($\textnormal{d}$B) values of denoising results for four competing state-of-the-art image denosing methods. Top Left: BM3D \cite{10}; Top Right: EPLL \cite{44}; Bottom Left: GSRC-BM3D; Bottom Right: GSRC-EPLL.}
\centering  
\begin{tabular}{||c||c|c||c|c||c|c||c|c||c|c||c|c||c|c|c|}
\hline
 \multirow{1}{*}{$\sigma$}&\multicolumn{2}{|c||}{20}&\multicolumn{2}{|c||}{30}&\multicolumn{2}{|c||}{40}
 &\multicolumn{2}{|c||}{50}&\multicolumn{2}{|c||}{75}&\multicolumn{2}{|c||}{100}\\
 \hline
  \hline
   \hline
 \multirow{2}{*}{Barbara}
 &  31.24      &  29.85      &  29.08      &  27.58      &  27.26     &  25.99      &  26.42      &  24.86      &  24.53      &  23.00      &  23.20   &  21.89\\
 \cline{2-13}
 &\textbf{31.65}  &\textbf{31.56}   &\textbf{29.50}  &\textbf{29.42}  &\textbf{27.92}  &\textbf{27.85} &\textbf{26.80}  &\textbf{26.54} &\textbf{24.80}  &\textbf{24.52}  &\textbf{23.57}   &\textbf{23.42} \\
 \hline \hline
 \multirow{2}{*}{Elaine}
 &  32.51      &  32.16      &  30.52      &  30.15      &  28.95     &  28.73      &  27.96      &  27.63      &  25.93      &  25.60      &  24.48   &  24.16\\
 \cline{2-13}
 &\textbf{32.63}  &\textbf{32.65}   &\textbf{30.61}  &\textbf{30.66}  &\textbf{29.09}  &\textbf{29.14} &\textbf{28.07}  &\textbf{28.02} &\textbf{26.14}  &\textbf{26.11}  &\textbf{24.74}   &\textbf{24.67} \\
 \hline \hline
  \multirow{2}{*}{flower}
 &  30.01      &  30.01      &  27.97      &  27.95      &  26.48     &  26.55      &  25.49      &  25.51      &  23.82      &  23.59      &  22.66   &  22.39\\
 \cline{2-13}
 &\textbf{30.47}  &\textbf{30.41}   &\textbf{28.28}  &\textbf{28.32}  &\textbf{26.92}  &\textbf{26.93} &\textbf{25.92}  &\textbf{25.89} &\textbf{24.17}  &\textbf{24.16}  &\textbf{22.94}   &\textbf{22.89} \\
 \hline \hline
 \multirow{2}{*}{foreman}
 &  34.54      &  33.67      &  32.75      &  31.70      &  31.29     &  30.28      &  30.36      &  29.20      &  28.07      &  27.24      &  26.51   &  25.91\\
 \cline{2-13}
 &\textbf{34.81}  &\textbf{34.82}   &\textbf{33.34}  &\textbf{33.34}  &\textbf{32.00}  &\textbf{32.13} &\textbf{31.06}  &\textbf{31.04} &\textbf{29.12}  &\textbf{29.12}  &\textbf{27.75}   &\textbf{27.77} \\
 \hline \hline
 \multirow{2}{*}{Hill}
 &  30.20      &  30.12      &  28.41      &  28.28      &  27.11     &  27.04      &  26.28      &  26.10      &  24.71      &  24.50     &  23.62   &  23.47\\
 \cline{2-13}
 &\textbf{30.29}  &\textbf{30.28}   &\textbf{28.45}  &\textbf{28.47}  &\textbf{27.24}  &\textbf{27.23} &\textbf{26.35}  &\textbf{26.29} &\textbf{24.82}  &\textbf{24.78}  &\textbf{23.82}   &\textbf{23.78} \\
 \hline \hline
 \multirow{2}{*}{House}
 &  33.77      &  32.99      &  32.09      &  31.24      &  30.65     &  29.89      &  29.69      &  28.79      &  27.51      &  26.70     &  25.87   &  25.21\\
 \cline{2-13}
 &\textbf{34.08}  &\textbf{34.03}   &\textbf{32.63}  &\textbf{32.54}  &\textbf{31.47}  &\textbf{31.45} &\textbf{30.43}  &\textbf{30.45} &\textbf{28.48}  &\textbf{28.53}  &\textbf{26.95}   &\textbf{26.97} \\
 \hline \hline
 \multirow{2}{*}{lena}
 &  31.52      &  31.25      &  29.46      &  29.18      &  27.82     &  27.78      &  26.90      &  26.68      &  25.17      &  24.75     &  23.87   &  23.46\\
 \cline{2-13}
 &\textbf{31.87}  &\textbf{31.83}   &\textbf{29.76}  &\textbf{29.78}  &\textbf{28.31}  &\textbf{28.30} &\textbf{27.31}  &\textbf{27.22} &\textbf{25.56}  &\textbf{25.52}  &\textbf{24.39}   &\textbf{24.41} \\
 \hline \hline
  \multirow{2}{*}{lin}
 &  32.83      &  32.62      &  30.95      &  30.67      &  29.52     &  29.32      &  28.71      &  28.26      &  26.96      &  26.36     &  26.00   &  25.05\\
 \cline{2-13}
 &\textbf{33.06}  &\textbf{33.03}   &\textbf{31.18}  &\textbf{31.13}  &\textbf{29.85}  &\textbf{29.86} &\textbf{28.93}  &\textbf{28.90} &\textbf{27.19}  &\textbf{27.17}  &\textbf{25.99}   &\textbf{25.97} \\
 \hline \hline
 \multirow{2}{*}{Monarch}
 &  30.35      &  30.49      &  28.36      &  28.36      &  26.72     &  26.89      &  25.82      &  25.78      &  23.91      &  23.73     &  22.52   &  22.24\\
 \cline{2-13}
 &\textbf{31.07}  &\textbf{31.10}   &\textbf{28.83}  &\textbf{28.88}  &\textbf{27.43}  &\textbf{27.45} &\textbf{26.38}  &\textbf{26.41} &\textbf{24.43}  &\textbf{24.49}  &\textbf{23.06}   &\textbf{23.09} \\
 \hline \hline
  \multirow{2}{*}{Parrot}
 &  32.32      &  32.00      &  30.33      &  30.00      &  28.64     &  28.60      &  27.88      &  27.53      &  25.94      &  25.56     &  24.60   &  24.08\\
 \cline{2-13}
 &\textbf{32.64}  &\textbf{32.58}   &\textbf{30.75}  &\textbf{30.68}  &\textbf{29.38}  &\textbf{29.33} &\textbf{28.35}  &\textbf{28.29} &\textbf{26.33}  &\textbf{26.31}  &\textbf{24.94}   &\textbf{24.96} \\
 \hline \hline
 \multirow{2}{*}{pentagon}
 &  28.23      &  27.96      &  26.41      &  26.06      &  25.10     &  24.79      &  24.21      &  23.83      &  22.59      &  22.18     &  21.45   &  21.12\\
 \cline{2-13}
 &\textbf{28.61}  &\textbf{28.57}   &\textbf{26.63}  &\textbf{26.60}  &\textbf{25.35}  &\textbf{25.30} &\textbf{24.42}  &\textbf{24.39} &\textbf{22.89}  &\textbf{22.83}  &\textbf{21.81}   &\textbf{21.73} \\
 \hline \hline
  \multirow{2}{*}{peppers}
 &  30.49      &  30.46      &  28.66      &  28.66      &  27.26     &  27.35      &  26.17      &  26.35      &  24.43      &  24.49     &  23.17   &  23.25\\
 \cline{2-13}
 &\textbf{30.72}  &\textbf{30.72}   &\textbf{28.81}  &\textbf{28.84}  &\textbf{27.51}  &\textbf{27.56} &\textbf{26.53}  &\textbf{26.56} &\textbf{24.67}  &\textbf{24.70}  &\textbf{23.44}   &\textbf{23.46} \\
 \hline \hline
 \multirow{2}{*}{\textbf{Average}}
 &  31.50      &  31.13      &  29.58      &  29.15      &  28.07     &  27.77      &  27.16      &  26.71      &  25.30      &  24.81     &  23.96   &  23.52\\
 \cline{2-13}
 &\textbf{31.83}  &\textbf{31.80}   &\textbf{29.90}  &\textbf{29.89}  &\textbf{28.54}  &\textbf{28.54} &\textbf{27.55}  &\textbf{27.50} &\textbf{25.72}  &\textbf{25.69}  &\textbf{24.45}   &\textbf{24.43} \\
 \hline \hline
\end{tabular}
\label{lab:2}
\end{table*}
\begin{figure*}[!htbp]
\begin{minipage}[b]{1\linewidth}
\centering
\centerline{\includegraphics[width=18cm]{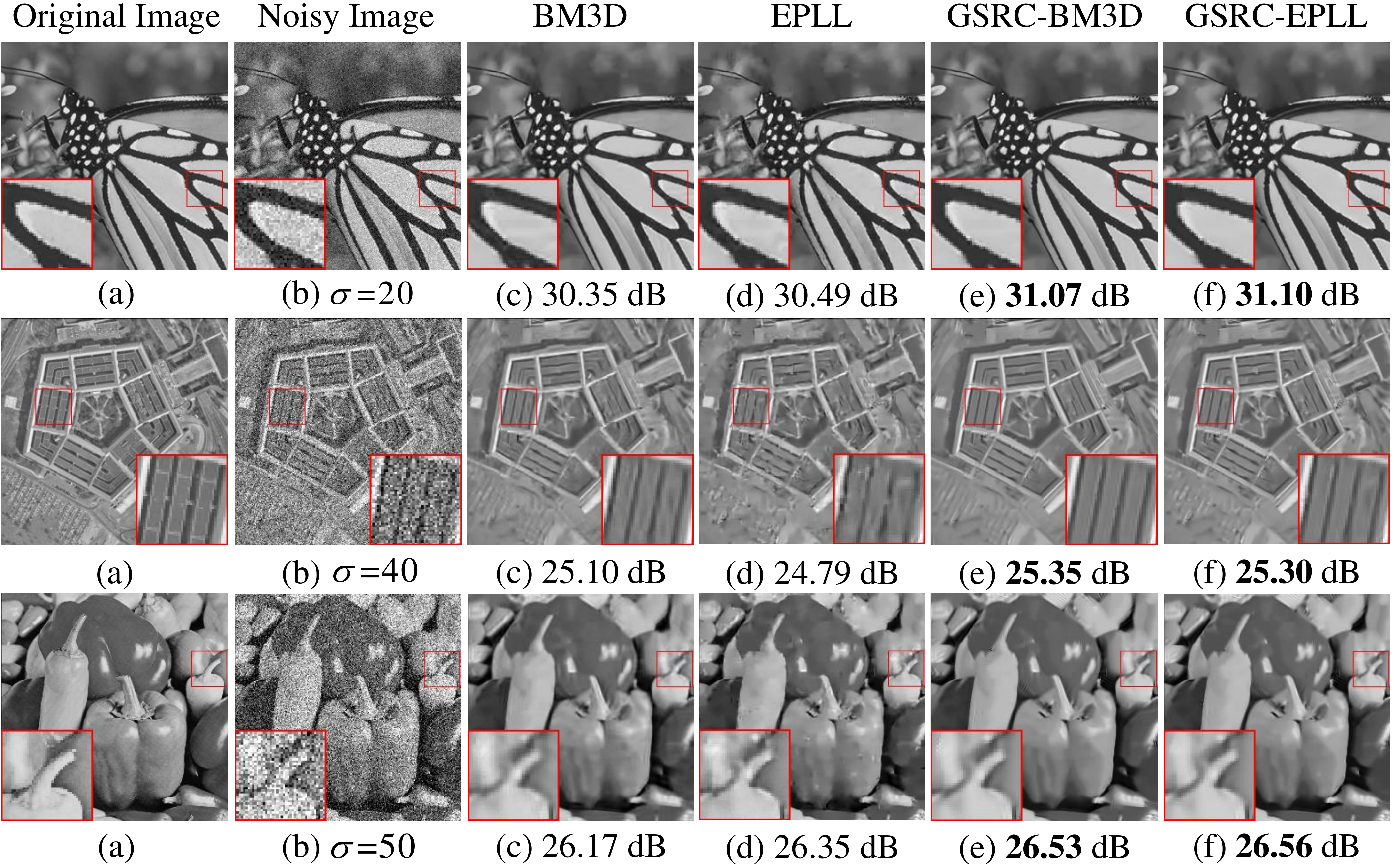}}
\end{minipage}
\caption{Denoising results of BM3D, EPLL, GSRC-BM3D and GSRC-EPLL on test image $\emph{Monarch}, \emph{pentagon}$ and $\emph{peppers}$ with $\sigma=20, 40$ and 50, respectively. (a) Original image; (b) Noisy Image; (c) Pre-filtering BM3D \cite{10} results. (d) Pre-filtering EPLL \cite{44} results; (e) GSRC-BM3D results; (f) GSRC-EPLL results.}
\label{fig:7}
\end{figure*}
\begin{table*}[!htbp]
\caption{PSNR ($\textnormal{d}$B) Comparison of NCSR \cite{19}, GID \cite{52}, LINC \cite{53}, MS-EPLL \cite{54}, AST-NLS \cite{55}, WNNM \cite{27}, GSRC-BM3D and GSRC-EPLL.}
\scriptsize
\centering  
\begin{tabular}{|c|c|c|c|c|c|c|c|c||c|c|c|c|c|c|c|c|c|}
\hline
  \multicolumn{1}{|c|}{}&\multicolumn{8}{|c||}{$\sigma=20$}&\multicolumn{8}{|c|}{$\sigma=30$}\\
\hline
\multirow{2}{*}{\textbf{{}}}&\multirow{2}{*}{\textbf{{NCSR}}}&\multirow{2}{*}{\textbf{{GID}}}&\multirow{2}{*}{\textbf{{LINC}}}
&{\textbf{{MS-}}}&{\textbf{{AST-}}}&\multirow{2}{*}{\textbf{{WNNM}}}&{\textbf{{GSRC-}}}&{\textbf{{GSRC-}}}&\multirow{2}{*}{\textbf{{NCSR}}}
&\multirow{2}{*}{\textbf{{GID}}}&\multirow{2}{*}{\textbf{{LINC}}}&{\textbf{{MS-}}}&{\textbf{{AST-}}}&\multirow{2}{*}{\textbf{{WNNM}}}
&{\textbf{{GSRC-}}}&{\textbf{{GSRC-}}}\\
& & & &{\textbf{EPLL}}&{\textbf{NLS}}& &{\textbf{BM3D}}&{\textbf{EPLL}}& & & &{\textbf{EPLL}}&{\textbf{NLS}}& &{\textbf{BM3D}}&{\textbf{EPLL}}\\
\hline
 \multirow{1}{*}{Barbara}  &    31.10     &  29.81      & \textbf{31.70}      & 30.07      & 31.43      & 31.60     & 31.65     & 31.56
                           &    28.68     &  27.35      & 29.53      & 27.72      & 29.13      & \textbf{29.68}     & 29.50     & 29.42\\
\hline
 \multirow{1}{*}{Elaine}   &    32.39     &  31.23      & 32.59      & 32.50      & 32.45      & 32.55     & \textbf{32.63}     & \textbf{32.65}
                           &    30.25     &  28.98      & 30.39      & 30.53      & 30.32      & \textbf{30.75}     & 30.61     & 30.66\\
\hline
 \multirow{1}{*}{flower}   &    30.05     &  29.12      & 30.30      & 30.10      & 30.28      & 30.34     & \textbf{30.47}     & \textbf{30.41}
                           &    27.86     &  27.01      & 28.13      & 28.05      & 28.20      & 28.26     & \textbf{28.29}     & \textbf{28.32}\\
\hline
 \multirow{1}{*}{foreman}  &    34.42     &  33.08      & 34.76      & 34.09      & 34.55      & 34.73     & \textbf{34.81}     & \textbf{34.82}
                           &    32.61     &  30.92      & 32.93      & 32.34      & 32.79      & 33.00     & \textbf{33.34}     & \textbf{33.34}\\
\hline
 \multirow{1}{*}{Hill}     &    30.03     &  29.06      & 30.14      & 30.20      & 30.22      & 30.27     & \textbf{30.29}     & \textbf{30.29}
                           &    28.15     &  27.05      & 28.29      & 28.41      & 28.37      & 28.41     & \textbf{28.45}     & \textbf{28.47}\\
\hline
 \multirow{1}{*}{House}    &    33.81     &  32.68      & 33.82      & 33.27      & 33.87      & 34.01     & \textbf{34.08}     & \textbf{34.03}
                           &    32.01     &  30.50      & 32.26      & 31.71      & 32.26      & 32.52     & \textbf{32.63}     & \textbf{32.54}\\
\hline
 \multirow{1}{*}{lena}     &    31.48     &  30.33      & 31.80      & 31.48      & 31.63      & 31.72     & \textbf{31.87}     & \textbf{31.83}
                           &    29.32     &  28.36      & \textbf{29.82}      & 29.46     & 29.50      & 29.45     & 29.76     & 29.78\\
\hline
 \multirow{1}{*}{lin}      &    32.66     &  31.74      & \textbf{33.04}      & 32.80      & 32.84      & 33.00     & \textbf{33.06}     & 33.03
                           &    30.65     &  29.63      & 31.03      & 30.96     & 30.83      & 31.07     & \textbf{31.18}     & \textbf{31.13}\\
\hline
 \multirow{1}{*}{Monarch}  &    30.52     &  29.75      & 30.64      & 30.59      & 30.84      & \textbf{31.10}     & 31.07     & \textbf{31.11}
                           &    28.38     &  27.68      & 28.74      & 28.49      & 28.70      & \textbf{28.91}     & 28.83     & 28.88\\
\hline
 \multirow{1}{*}{Parrot}   &    32.25     &  31.24      & 32.50      & 32.21      & 32.42      & \textbf{32.66}     & 32.64     & 32.58
                           &    30.20     &  29.33      & 30.64      & 30.29      & 30.37      & \textbf{30.78}     & 30.75     & 30.68\\
\hline
 \multirow{1}{*}{pentagon} &    28.27     &  27.39      & 28.43      & 27.99      & 28.49      & 28.49     & \textbf{28.61}     & \textbf{28.57}
                           &    26.27     &  25.32      & 26.42      & 26.05      & 26.57      & \textbf{26.67}     & 26.63     & 26.60\\
\hline
 \multirow{1}{*}{peppers}   &    30.35     &  29.78      & 30.50      & 30.60      & 30.61      & 30.70     & \textbf{30.72}     & \textbf{30.72}
                           &    28.41     &  27.86      & 28.79      & \textbf{28.85}      & 28.75      & 28.84     & 28.81     & 28.84\\
\hline
 \multirow{1}{*}{\textbf{Average}} &    31.45     &  30.43      & 31.69      & 31.32      & 31.64      & 31.76     & \textbf{31.83}     & \textbf{31.80}
                           &    29.40     &  28.33      & 29.75      & 29.41      & 29.65      & 29.86     & \textbf{29.90}     & \textbf{29.89}\\
\hline
  \multicolumn{1}{|c|}{}&\multicolumn{8}{|c||}{$\sigma=40$}&\multicolumn{8}{|c|}{$\sigma=50$}\\
\hline
\multirow{2}{*}{\textbf{{}}}&\multirow{2}{*}{\textbf{{NCSR}}}&\multirow{2}{*}{\textbf{{GID}}}&\multirow{2}{*}{\textbf{{LINC}}}
&{\textbf{{MS-}}}&{\textbf{{AST-}}}&\multirow{2}{*}{\textbf{{WNNM}}}&{\textbf{{GSRC-}}}&{\textbf{{GSRC-}}}&\multirow{2}{*}{\textbf{{NCSR}}}
&\multirow{2}{*}{\textbf{{GID}}}&\multirow{2}{*}{\textbf{{LINC}}}&{\textbf{{MS-}}}&{\textbf{{AST-}}}&\multirow{2}{*}{\textbf{{WNNM}}}
&{\textbf{{GSRC-}}}&{\textbf{{GSRC-}}}\\
& & & &{\textbf{EPLL}}&{\textbf{NLS}}& &{\textbf{BM3D}}&{\textbf{EPLL}}& & & &{\textbf{EPLL}}&{\textbf{NLS}}& &{\textbf{BM3D}}&{\textbf{EPLL}}\\
\hline
 \multirow{1}{*}{Barbara}  &    27.25     &  25.77      & 27.77      & 26.05      & 27.41      & \textbf{27.86}     & \textbf{27.92}     & 27.85
                           &    26.13     &  24.52      & 26.27      & 25.06      & 26.43      & \textbf{26.83}     & 26.80     & 26.54\\
\hline
 \multirow{1}{*}{Elaine}   &    28.91     &  27.67      & 28.96      & \textbf{29.13}      & 28.69      & 29.05     & 29.09     & \textbf{29.14}
                           &    27.68     &  26.54      & 27.73      & 27.98      & 27.68      & \textbf{28.08}     & 28.07     & 28.02\\
\hline
 \multirow{1}{*}{flower}   &    26.35     &  25.60      & 26.79      & 26.64      & 26.75      & 26.85     & \textbf{26.92}     & \textbf{26.93}
                           &    25.31     &  24.42      & 25.47      & 25.56      & 25.77      & 25.80     & \textbf{25.92}     & \textbf{25.89}\\
\hline
 \multirow{1}{*}{foreman}  &    31.52     &  29.61      & 31.31      & 31.05      & 31.29      & 31.54     & \textbf{32.00}     & \textbf{32.13}
                           &    30.41     &  28.64      & 30.33      & 30.04      & 30.46      & 30.75     & \textbf{31.06}     & \textbf{31.04}\\
 \hline
 \multirow{1}{*}{Hill}     &    26.91     &  25.87      & 26.84      & 27.18      & 27.05      & \textbf{27.29}     & 27.24     & 27.23
                           &    26.01     &  25.05      & 26.03      & 26.28      & 26.16      & 26.14     & \textbf{26.35}     & \textbf{26.29}\\
 \hline
 \multirow{1}{*}{House}    &    30.79     &  29.02      & 31.00      & 30.47      & 30.91      & 31.31     & \textbf{31.47}     & \textbf{31.46}
                           &    29.61     &  27.76      & 29.87      & 29.47      & 30.13      & 30.32     & \textbf{30.43}     & \textbf{30.45}\\
  \hline
 \multirow{1}{*}{lena}     &    28.00     &  26.98      & 28.13      & 28.05      & 28.00      & \textbf{28.43}     & 28.31     & 28.30
                           &    26.94     &  25.82      & 26.94      & 26.97      & 27.08      & \textbf{27.26}     & \textbf{27.31}     & 27.22\\
\hline
 \multirow{1}{*}{lin}      &    29.27     &  28.44      & \textbf{29.94}      & 29.68      & 29.39      & 29.78     & 29.85     & 29.86
                           &    28.23     &  27.50      & 28.85      & 28.69      & 28.50      & 28.83     & \textbf{28.93}     & \textbf{28.90}\\
 \hline
 \multirow{1}{*}{Monarch}  &    26.81     &  26.32      & 27.14      & 27.06      & 27.20      & \textbf{27.47}     & 27.43     & 27.45
                           &    25.73     &  25.28      & 25.88      & 25.93      & 26.12      & 26.18     & \textbf{26.38}     & \textbf{26.41}\\
 \hline
 \multirow{1}{*}{Parrot}   &    28.77     &  28.01      & 29.26      & 28.94      & 28.87      & \textbf{29.33}     & \textbf{29.38}     & 29.33
                           &    27.67     &  26.79      & 28.23      & 27.90      & 27.92      & 28.15     & \textbf{28.35}     & \textbf{28.29}\\
 \hline
 \multirow{1}{*}{pentagon} &    24.93     &  23.95      & 24.96      & 24.75      & 25.22      & \textbf{25.41}     & 25.35     & 25.30
                           &    23.94     &  22.81      & 23.85      & 23.81      & 24.31      & \textbf{24.46}     & 24.42     & 24.39\\
  \hline
 \multirow{1}{*}{peppers}   &    27.09     &  26.47      & 27.39      & 27.57      & 27.37      & \textbf{27.70}     & 27.51     & 27.56
                           &    26.02     &  25.48      & 26.47      & 26.55      & 26.36      & \textbf{26.56}     & 26.53     & \textbf{26.56}\\
\hline
 \multirow{1}{*}{\textbf{Average}}   &    28.05     &  26.97      & 28.29      & 28.05      & 28.18      & 28.50     & \textbf{28.54}     & \textbf{28.55}
                           &    26.97     &  25.88      & 27.16      & 27.02      & 27.24      & 27.44     & \textbf{27.55}     & \textbf{27.50}\\
 \hline
\multicolumn{1}{|c|}{}&\multicolumn{8}{|c||}{$\sigma=75$}&\multicolumn{8}{|c|}{$\sigma=100$}\\
\hline
\multirow{2}{*}{\textbf{{}}}&\multirow{2}{*}{\textbf{{NCSR}}}&\multirow{2}{*}{\textbf{{GID}}}&\multirow{2}{*}{\textbf{{LINC}}}
&{\textbf{{MS-}}}&{\textbf{{AST-}}}&\multirow{2}{*}{\textbf{{WNNM}}}&{\textbf{{GSRC-}}}&{\textbf{{GSRC-}}}&\multirow{2}{*}{\textbf{{NCSR}}}
&\multirow{2}{*}{\textbf{{GID}}}&\multirow{2}{*}{\textbf{{LINC}}}&{\textbf{{MS-}}}&{\textbf{{AST-}}}&\multirow{2}{*}{\textbf{{WNNM}}}
&{\textbf{{GSRC-}}}&{\textbf{{GSRC-}}}\\
& & & &{\textbf{EPLL}}&{\textbf{NLS}}& &{\textbf{BM3D}}&{\textbf{EPLL}}& & & &{\textbf{EPLL}}&{\textbf{NLS}}& &{\textbf{BM3D}}&{\textbf{EPLL}}\\
\hline
 \multirow{1}{*}{Barbara}  &    24.06     &  22.43      & 24.04      & 23.19      & 24.40      & \textbf{24.79}     & \textbf{24.80}     & 24.52
                           &    22.70     &  21.40      & 22.39      & 22.11      & 23.19      & 23.26     & \textbf{23.57}     & \textbf{23.42}\\
  \hline
 \multirow{1}{*}{Elaine}   &    25.34     &  24.54      & 25.42      & 25.99      & 25.51      & 25.94     & \textbf{26.14}     & \textbf{26.11}
                           &    23.77     &  23.21      & 23.92      & 24.52      & 23.94      & 24.54     & \textbf{24.74}     & \textbf{24.67}\\
  \hline
 \multirow{1}{*}{flower}   &    23.50     &  22.72      & 23.30      & 23.68      & 23.87      & 23.88     & \textbf{24.17}     & \textbf{24.16}
                           &    22.22     &  20.69      & 21.96      & 22.50      & 22.50      & 22.70     & \textbf{22.94}     & \textbf{22.89}\\
  \hline
  \multirow{1}{*}{foreman} &    28.18     &  26.71      & 28.11      & 28.14      & 28.54      & 28.48     & \textbf{29.12}     & \textbf{29.12}
                           &    26.55     &  25.33      & 26.55      & 26.84      & 27.32      & 27.39     & \textbf{27.75}     & \textbf{27.77}\\
  \hline
  \multirow{1}{*}{Hill}    &    24.43     &  23.62      & 24.13      & 24.72      & 24.42      & 24.70     & \textbf{24.82}     & \textbf{24.79}
                           &    23.27     &  22.75      & 23.21      & 23.74      & 23.33      & 23.63     & \textbf{23.82}     & \textbf{23.78}\\
  \hline
  \multirow{1}{*}{House}   &    27.16     &  25.23      & 27.56      & 27.45      & 28.06      & 28.25     & \textbf{28.48}     & \textbf{28.53}
                           &    25.49     &  22.38      & 26.11      & 25.99      & 26.52      & 26.68     & \textbf{26.95}     & \textbf{26.97}\\
  \hline
   \multirow{1}{*}{lena}   &    25.02     &  23.78      & 25.12      & 25.11      & 25.32      & 25.38     & \textbf{25.56}     & \textbf{25.52}
                           &    23.63     &  22.43      & 23.67      & 23.91      & 24.17      & 24.08     & \textbf{24.39}     & \textbf{24.41}\\
  \hline
\multirow{1}{*}{lin}       &    26.22     &  25.50      & 26.86      & 26.91      & 26.72      & 26.94     & \textbf{27.19}     & \textbf{27.17}
                           &    24.85     &  24.14      & 25.23      & 25.63      & 25.42      & 25.67     & \textbf{25.99}     & \textbf{25.97}\\
  \hline
\multirow{1}{*}{Monarch}   &    23.67     &  22.77      & 23.91      & 23.92      & 24.11      & 24.31     & \textbf{24.43}     & \textbf{24.49}
                           &    22.10     &  20.73      & 22.13      & 22.44      & 22.68      & 22.95     & \textbf{23.06}     & \textbf{23.09}\\
  \hline
\multirow{1}{*}{Parrot}    &    25.45     &  24.87      & 26.20      & 25.92      & 25.93      & \textbf{26.32}     & \textbf{26.33}     & 26.31
                           &    23.94     &  23.54      & 24.48      & 24.38      & 24.61      & 24.85     & \textbf{24.94}     & \textbf{24.96}\\
  \hline
\multirow{1}{*}{pentagon}  &    22.04     &  21.31      & 22.26      & 22.19      & 22.49      & 22.65     & \textbf{22.89}     & \textbf{22.83}
                           &    20.92     &  19.53      & 21.05      & 21.15      & 21.24      & 21.56     & \textbf{21.81}     & \textbf{21.73}\\
  \hline
 \multirow{1}{*}{peppers}  &    24.07     &  23.44      & 24.44      & \textbf{24.74}      & 24.47      & 24.61     & 24.67     & 24.70
                           &    22.68     &  22.09      & 22.83      & \textbf{23.51}      & 23.18      & 23.19     & 23.44     & 23.46\\
  \hline
   \multirow{1}{*}{\textbf{Average}}  &    24.93     &  23.91      & 25.11      & 25.16      & 25.32      & 25.52     & \textbf{25.72}     & \textbf{25.69}
                           &    23.51     &  22.35     & 23.63     & 23.89      & 24.01      & 24.21     & \textbf{24.45}     & \textbf{24.43}\\
  \hline
\end{tabular}
\label{lab:3}
\end{table*}
\begin{figure*}[!htbp]
\begin{minipage}[b]{1\linewidth}
\centering
\centerline{\includegraphics[width=18cm]{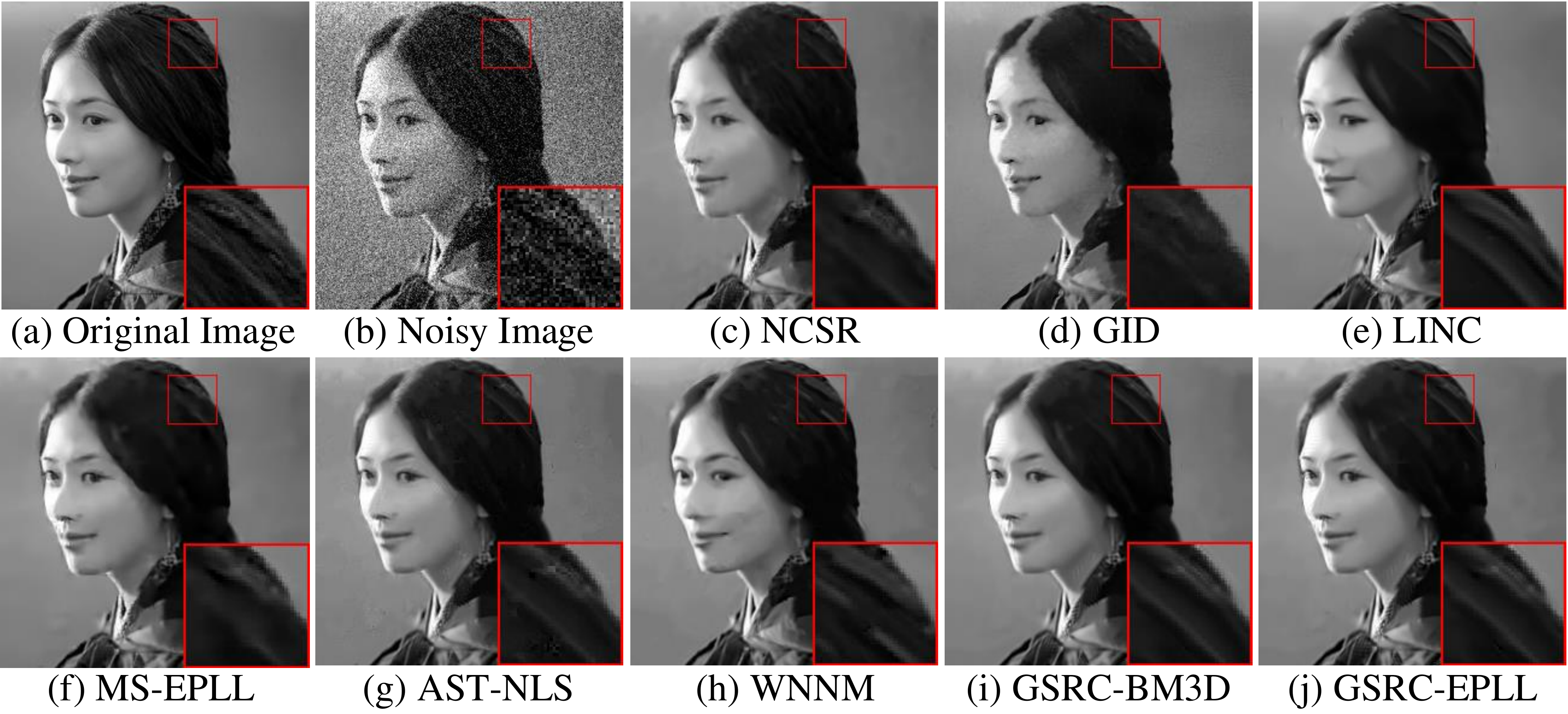}}
\end{minipage}
\caption{Denoising images of \emph{lin} by different methods ($\sigma=30$). (a) Original image; (b) Noisy image; (c) NCSR \cite{19} (PSNR= 30.65dB, SSIM=0.8632); (d) GID \cite{52} (PSNR= 29.63dB, SSIM=0.8287); (e) LINC \cite{53} (PSNR= 31.03dB, SSIM=0.8670); (f) MS-EPLL \cite{54} (PSNR= 30.96dB, SSIM=0.8688); (g) AST-NLS \cite{55} (PSNR= 30.83dB, SSIM=0.8465);  (h) WNNM \cite{27} (PSNR= 31.07dB, SSIM=0.8657);  (i) GSRC-BM3D (PSNR= \textbf{31.18dB}, SSIM =\textbf{0.8743}); (j) GSRC-EPLL (PSNR = \textbf{31.13dB}, SSIM=\textbf{0.8704}).}
\label{fig:8}
\end{figure*}
\subsection{Performance Comparison with the State-of-the-Art Methods}
In this subsection, we validate the performance of the proposed GSRC and compare it with recently proposed state-of-the-art denoising methods, including BM3D \cite{10}, EPLL \cite{44}, NCSR \cite{19}, GID \cite{52}, LINC \cite{53}, MS-EPLL \cite{54}, AST-NLS \cite{55} and WNNM \cite{27}. For all the competing methods, the source codes are obtained from the original authors. We used the default parameters in their software packages.

First, we compare GSRC-BM3D, GSRC-EPLL with BM3D and EPLL method, respectively. In Table~\ref{lab:2}, we report the PSNR results for different noise variances for the 12 test images in Fig.~\ref{fig:6}. It can be seen that GSRC-BM3D, GSRC-EPLL are significantly better than BM3D and EPLL with an average gain of about 0.40dB and 0.79dB, respectively. The visual quality comparisons in the case of  $\sigma=20, 40$ and 50 for  test images \emph{Monarch}, \emph{pentagon} and \emph{peppers} are provided in Fig.~\ref{fig:7}, respectively. It can be found out that the over-smooth phenomena and undesirable artifacts are generated by BM3D and EPLL methods, respectively. In contrast, the proposed GSRC not only reduces most of the artifacts, but also provides better denoising performance on both edges and textures than BM3D and EPLL methods. Therefore,  these results validate the usefulness of the proposed GSRC model through the pre-filtering  BM3D and EPLL.

\begin{figure*}[!htbp]
\begin{minipage}[b]{1\linewidth}
\centering
\centerline{\includegraphics[width=18cm]{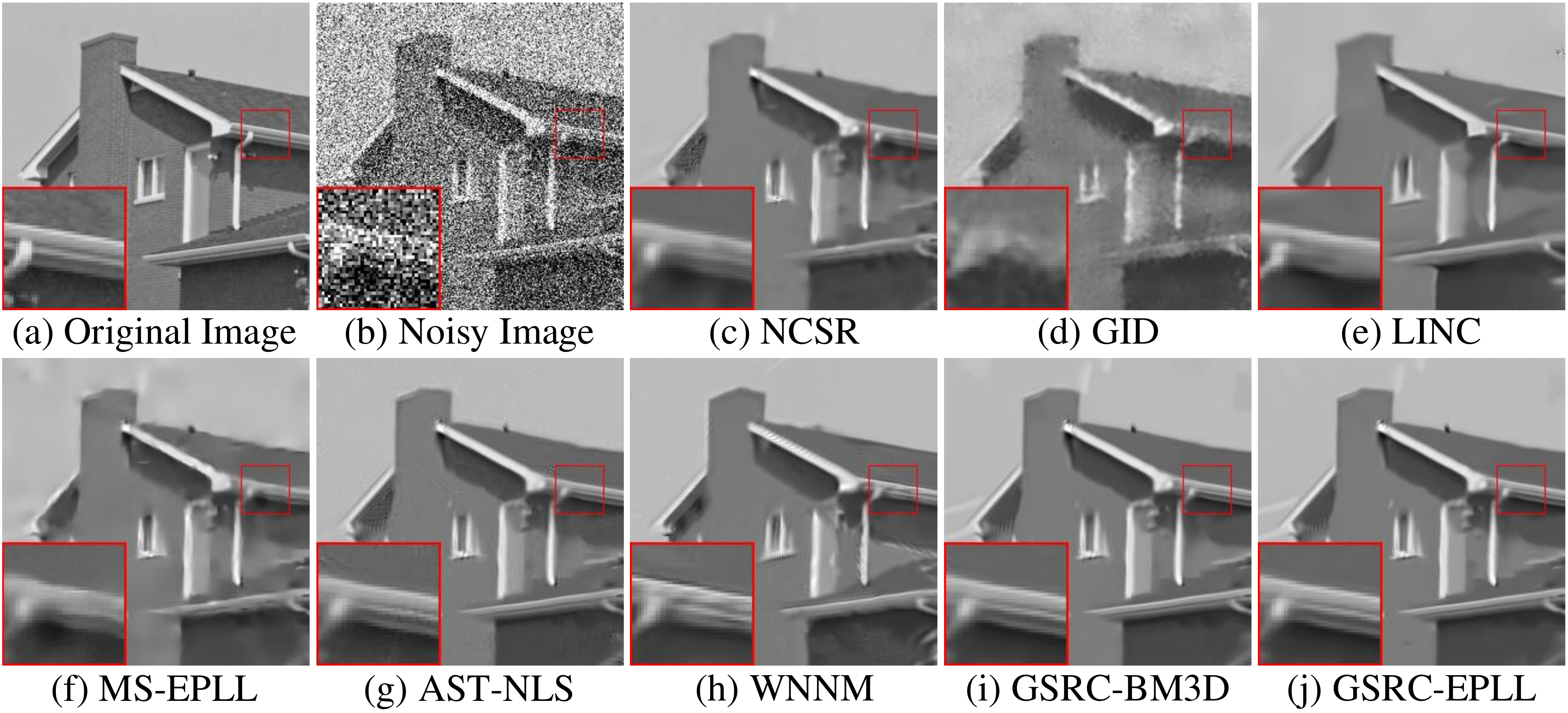}}
\end{minipage}
\caption{Denoising images of \emph{House} by different methods ($\sigma=75$). (a) Original image; (b) Noisy image; (c) NCSR \cite{19} (PSNR= 27.16dB, SSIM =0.7749); (d) GID \cite{52} (PSNR= 25.23dB, SSIM=0.7052); (e) LINC \cite{53} (PSNR= 27.56dB, SSIM=0.7850); (f) MS-EPLL \cite{54} (PSNR= 27.45dB, SSIM=0.7738); (g) AST-NLS \cite{55} (PSNR= 28.06dB, SSIM=0.7720);  (h) WNNM \cite{27} (PSNR= 28.25dB, SSIM=0.7883);  (i) GSRC-BM3D (PSNR= \textbf{28.48dB}, SSIM=\textbf{0.7992}); (j) GSRC-EPLL (PSNR = \textbf{28.53dB}, SSIM=\textbf{0.7998}).}
\label{fig:9}
\end{figure*}

Second, to further verify the performance of the proposed GSRC in image denoising, we compare it with six representative algorithms: NCSR \cite{19}, GID \cite{52}, LINC \cite{53}, MS-EPLL \cite{54}, AST-NLS \cite{55} and WNNM \cite{27}. Gaussian white noise with standard deviation $\sigma$=20, 30, 40, 50, 75 and 100 is added to the 12  test images. The PSNR results by the competing denoising methods are shown in Table~\ref{lab:3}. It can be seen that the proposed GSRC has achieved highly competitive denoising performance to other leading methods. Based on the pre-filtering BM3D \cite{10}, the proposed GSRC achieves 0.61dB, 1.68dB, 0.39dB, 0.52dB, 0.32dB and 0.11dB improvement on average over  NCSR, GID, LINC, MS-EPLL, AST-NLS  and WNNM, respectively. Meanwhile, based on the pre-filtering EPLL \cite{44}, the proposed GSRC achieves 0.59dB, 1.66dB, 0.37dB, 0.50dB, 0.30dB and 0.09dB improvement on average over  NCSR, GID, LINC, MS-EPLL, AST-NLS  and WNNM, respectively. The visual comparisons of the competing  methods at noise level 30 and 75 are shown in Fig.~\ref{fig:8}  and Fig.~\ref{fig:9}, respectively.
\begin{table}[!htbp]
\caption{Average PSNR ($\textnormal{d}$B) results of different denoising algorithms for Gaussian denoising with noise level 20, 30, 40, 50, 75 and 100 on BSD200 dataset \cite{51}.}
\centering  
\begin{tabular}{|c|c|c|c|c|c|c|}
\hline\hline
\multirow{1}{*}{\textbf{{$\sigma$}}}&{20}&{30}&{40}&{50}&{75}&{100}\\
 \hline
 \multirow{1}{*}{NCSR \cite{19}}        & 29.89 &  27.92 & 26.58 & 25.65 & 24.04 & 23.00 \\
 \hline
 \multirow{1}{*}{GID \cite{52}}         & 28.87 &  27.00 & 25.87 & 24.97 & 23.37 & 22.20 \\
 \hline
 \multirow{1}{*}{LINC \cite{53}}        & 29.92 &  27.94 & 26.61 & 25.64 & 23.98 & 22.91 \\
 \hline
 \multirow{1}{*}{MS-EPLL \cite{54}}     & 29.95 &  28.02 & 26.73 & 25.84 & 24.29 & 23.27 \\
 \hline
 \multirow{1}{*}{AST-NLS \cite{55}}     & 29.98 &  28.02 & 26.68 & 25.80 & 24.20 & 23.17 \\
 \hline
 \multirow{1}{*}{WNNM \cite{27}}        & \textbf{30.11} &  \textbf{28.17} & 26.88 & 25.96 & 24.42 & 23.37 \\
 \hline
 \multirow{1}{*}{GSRC-BM3D}     & \textbf{30.12} &  28.15 & \textbf{26.89} & \textbf{26.01} & \textbf{24.49} &\textbf{23.49} \\
 \hline
  \multirow{1}{*}{GSRC-EPLL}    & \textbf{30.11} &  \textbf{28.17} & \textbf{26.91} & \textbf{26.00} & \textbf{24.48} &\textbf{23.49} \\
 \hline
 \hline
\end{tabular}
\label{lab:4}
\end{table}
\begin{table}[!htbp]
\caption{Average PSNR ($\textnormal{d}$B) results of APS and No-APS scheme on 12 test images.}
\centering  
\begin{tabular}{|c|c|c|c|c|c|c|}
\hline\hline
  \multicolumn{1}{|c|}{Pre-filtering}&\multicolumn{6}{|c|}{BM3D}\\
\hline
\multirow{1}{*}{$\sigma$}&\multirow{1}{*}{{{20}}}&\multirow{1}{*}{{{30}}}&\multirow{1}{*}{{{40}}}
&{{{50}}}&{{{75}}}&\multirow{1}{*}{{{100}}}\\
\hline
 \multirow{1}{*}{No-APS}&    31.71      &  29.74      & 28.38     & 27.37    &  25.56     &  24.25    \\
\hline
 \multirow{1}{*}{APS}&    \textbf{31.83}     &  \textbf{29.90}      & \textbf{28.54}     & \textbf{27.55}    &  \textbf{25.72}     &  \textbf{24.45}    \\
\hline
  \multicolumn{1}{|c|}{Pre-filtering}&\multicolumn{6}{|c|}{EPLL}\\
\hline
\multirow{1}{*}{$\sigma$}&\multirow{1}{*}{{{20}}}&\multirow{1}{*}{{{30}}}&\multirow{1}{*}{{{40}}}
&{{{50}}}&{{{75}}}&\multirow{1}{*}{{{100}}}\\
\hline
 \multirow{1}{*}{No-APS}&    31.71      &  29.80     & 28.42     & 27.36    &  25.55    &  24.26    \\
\hline
 \multirow{1}{*}{APS}&    \textbf{31.80}     &  \textbf{29.89}      & \textbf{28.55}     & \textbf{27.50}    &  \textbf{25.69}     &  \textbf{24.43}    \\
\hline\hline
\end{tabular}
\label{lab:5}
\end{table}

\begin{figure*}[!htbp]
\begin{minipage}[b]{1\linewidth}
\centering
\centerline{\includegraphics[width=18cm]{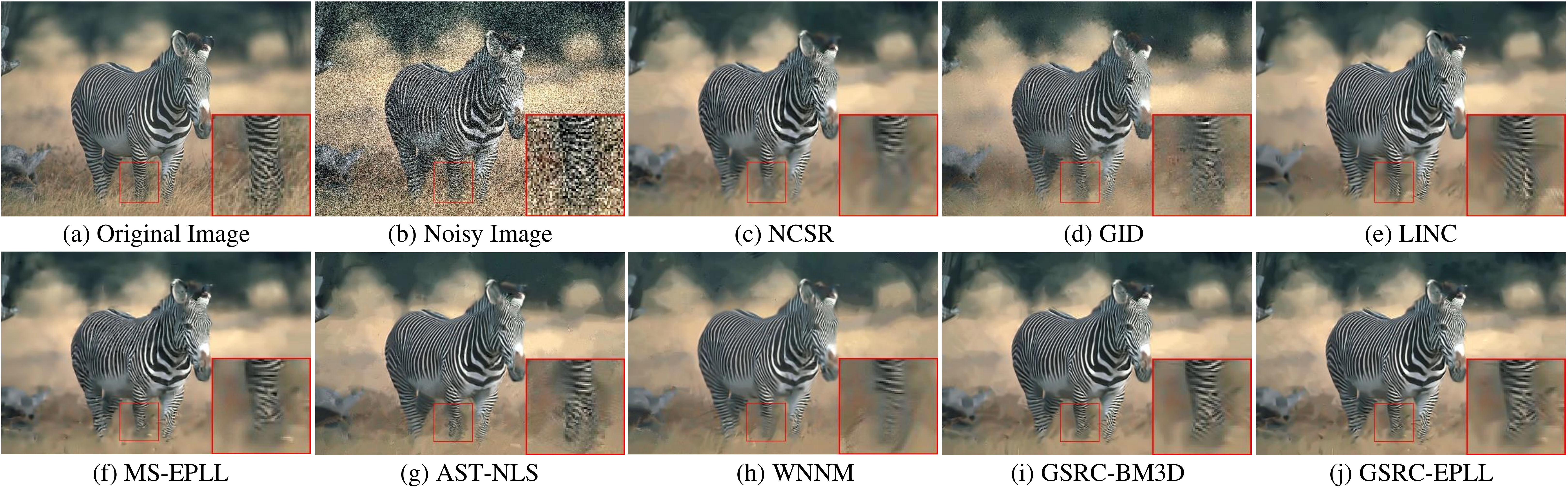}}
\end{minipage}
\caption{Denoising images of \emph{130066} by different methods ($\sigma=50$). (a) Original image; (b) Noisy image; (c) NCSR \cite{19} (PSNR= 25.69dB, SSIM=0.7800); (d) GID \cite{52} (PSNR= 25.40dB, SSIM=0.7152); (e) LINC \cite{53} (PSNR= 26.51dB, SSIM=0.7938); (f) MS-EPLL \cite{54} (PSNR= 25.67dB, SSIM=0.7833); (g) AST-NLS \cite{55} (PSNR= 26.27dB, SSIM=0.7715);  (h) WNNM \cite{27} (PSNR= 26.43dB, SSIM=0.7888);  (i) GSRC-BM3D (PSNR= \textbf{26.69dB}, SSIM=\textbf{0.7937}); (j) GSRC-EPLL (PSNR = \textbf{26.62dB}, SSIM=\textbf{0.7908}).}
\label{fig:10}
\end{figure*}

\begin{figure*}[!htbp]
\begin{minipage}[b]{1\linewidth}
\centering
\centerline{\includegraphics[width=18cm]{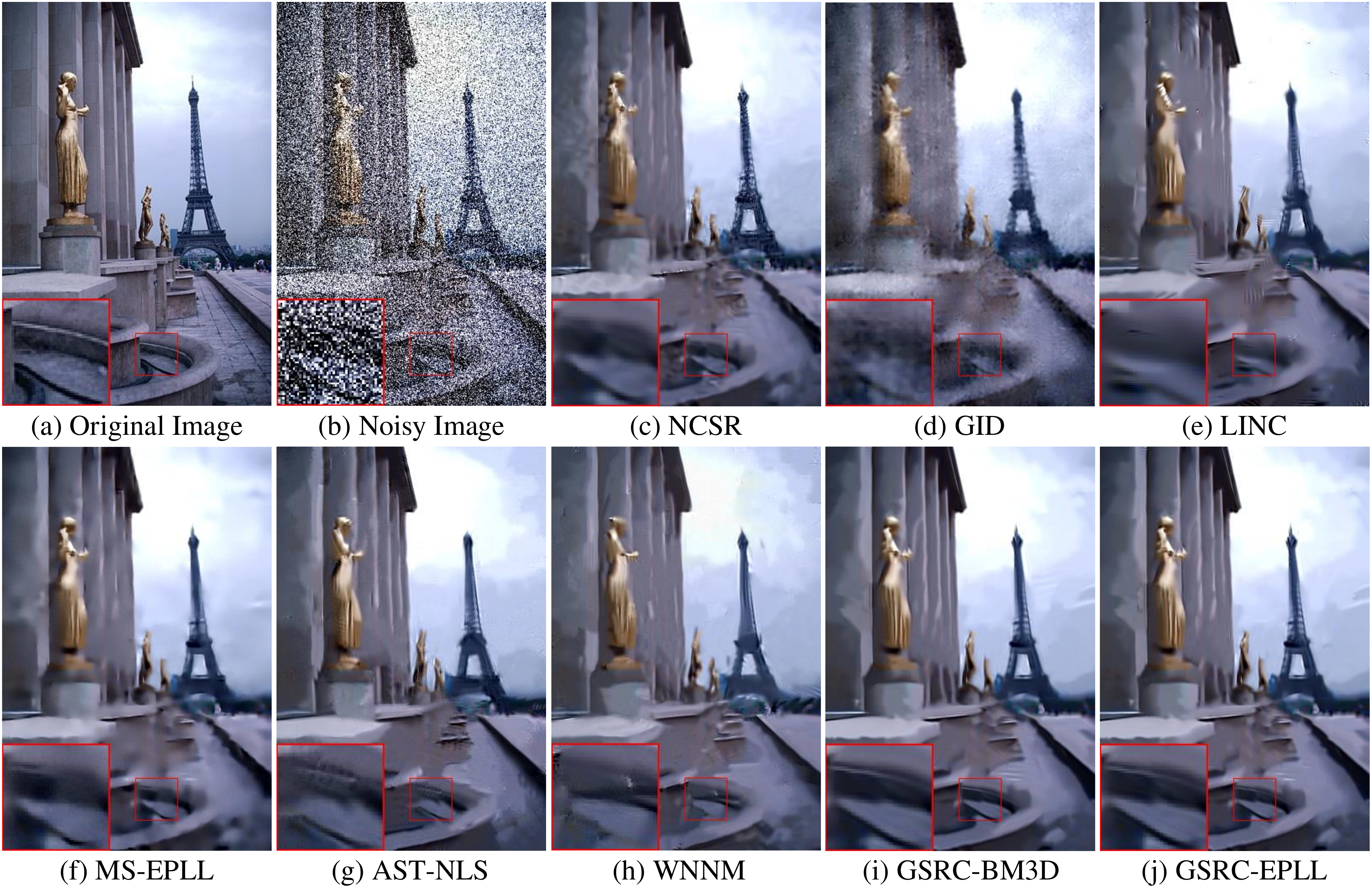}}
\end{minipage}
\caption{Denoising images of \emph{223004} by different methods ($\sigma=100$). (a) Original image; (b) Noisy image; (c) NCSR \cite{19} (PSNR= 23.73dB, SSIM=0.6882); (d) GID \cite{52} (PSNR= 22.66dB, SSIM=0.5997); (e) LINC \cite{53} (PSNR= 23.68dB, SSIM=0.6756); (f) MS-EPLL \cite{54} (PSNR= 24.21dB, SSIM=0.6869); (g) AST-NLS \cite{55} (PSNR= 24.24dB, SSIM=0.6861);  (h) WNNM \cite{27} (PSNR= 24.48dB, SSIM=0.7025);  (i) GSRC-BM3D (PSNR= \textbf{24.70dB}, SSIM =\textbf{0.7182}); (j) GSRC-EPLL (PSNR = \textbf{24.67dB}, SSIM=\textbf{0.7157}).}
\label{fig:11}
\end{figure*}

To further demonstrate our performance, we comprehensively evaluate the proposed GSRC on 200 test images from the BSD dataset \cite{51}. Table~\ref{lab:4} lists the average PSNR comparison results for a collection of  200 test images among eight competing methods at six noise levels  ($\sigma$=20, 30, 40, 50, 75 and 100). The visual comparisons of the denoising methods for test images $\emph{130066}$ and $\emph{223004}$ with $\sigma=50$ and 100 are shown in Fig.~\ref{fig:10} and Fig.~\ref{fig:11}, respectively. Obviously, one can observe that the proposed GSRC achieves very competitive denoising performance compared to WNNM.

\begin{figure*}[!htbp]
\begin{minipage}[b]{1\linewidth}
\centering
\centerline{\includegraphics[width=18cm]{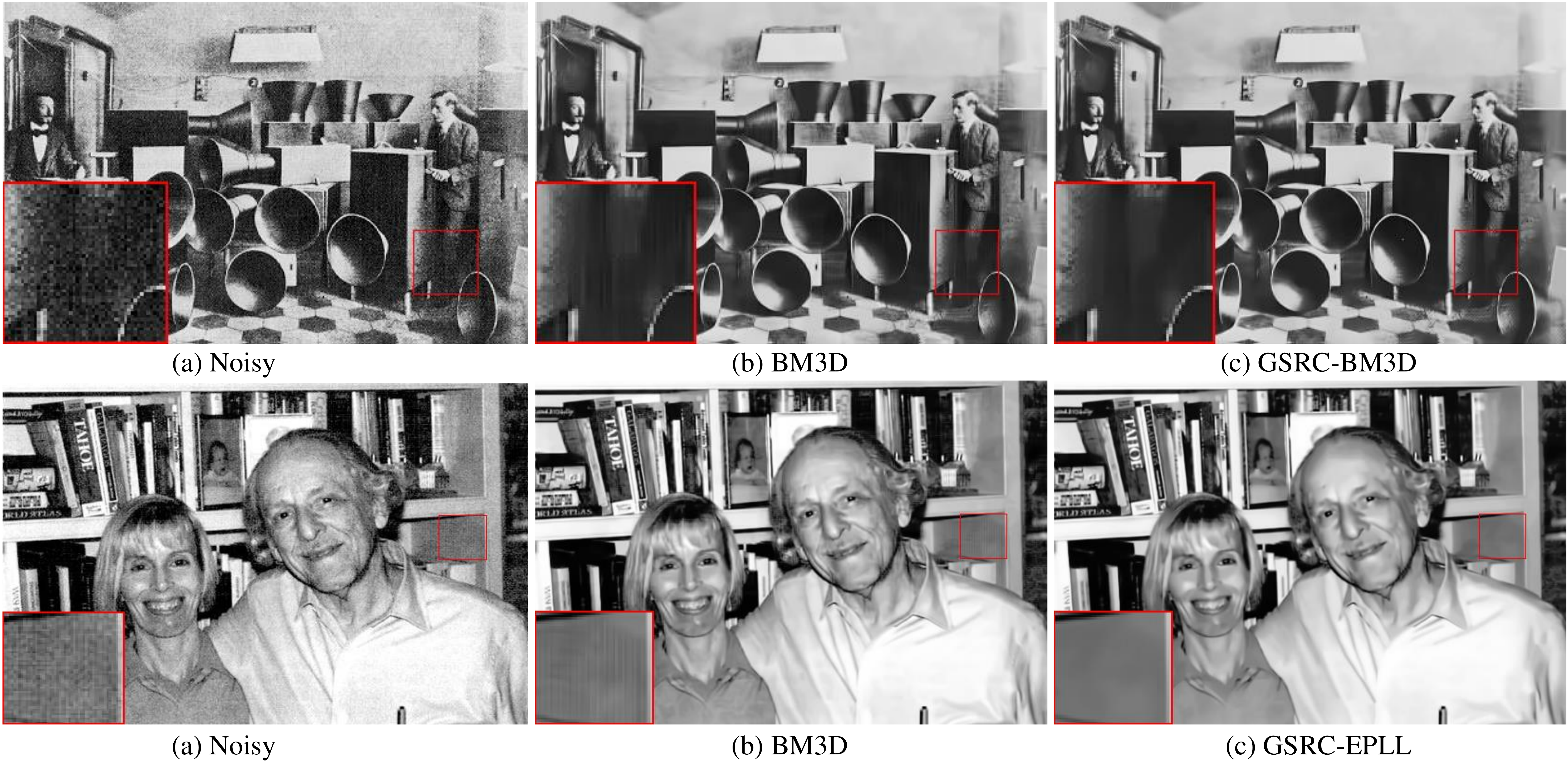}}
\end{minipage}
\caption{Visual comparisons of denoising results on real noisy images with unknown noise characteristics.}
\label{fig:12}
\end{figure*}

In addition, we apply the proposed GSRC to some real noisy images. Fig.~\ref{fig:12} shows the denoised images yielded by BM3D and our approach. It can be seen that the proposed GSRC can not only reduce the noise effectively, but also preserve the finer details. The results indicate the feasibility of  the proposed GSRC for some practical image denoising tasks.
\begin{table*}[!htbp]
\caption{Average run time ($\textnormal{s}$) on the 12 test images (size: $256\times 256$) with different methods.}
\centering  
\begin{tabular}{|c|c|c|c|c|c|c|c|c|}
\hline
\multirow{1}{*}{Methods}&{NCSR \cite{19}}&{GID \cite{52}}&{LINC \cite{53}}&{MS-EPLL \cite{54}}&{AST-NLS \cite{55}}&{WNNM \cite{27}}&{GSRC-BM3D}&{GSRC-EPLL}\\
 \hline
  \multirow{1}{*}{Average Time (s)}  & 348 &  346 & 257 & 191 &  459 & 202 & \textbf{72} & \textbf{148}\\
  \hline
\end{tabular}
\label{lab:6}
\end{table*}

To sum up, it can be easily found that  BM3D, EPLL, NCSR, GID, LINC, MS-EPLL, AST-NLS and WNNM still generate some undesirable artifacts and some details are lost. By contrast, the proposed GSRC is able to preserve the sharp edges and suppress undesirable artifacts more effectively than other competing methods. Such experimental findings clearly demonstrate that the GSRC  model is a stronger prior for the class of photographic images containing large variations in edges/textures.
\subsection{Comparison between APS and No-APS Scheme}
In this subsection, in order to  demonstrate the designed adaptive patch selection (APS) scheme effectively, we compare it with No-APS scheme. Table~\ref{lab:5} shows the average PSNR results of APS and No-APS schemes on 12 test images. It can be seen that the average PSNR results of APS scheme are better than No-APS. Therefore, the proposed APS scheme can boost the accuracy of nonlocal similar patch selection under the task of image denoising.
\subsection{Computational Cost}
Efficiency is another key factor in evaluating an algorithm. We then compare the speed of the proposed GSRC and six representative algorithms. All experiments are conducted under the Matlab 2012b environment on a machine with Intel (R) Core (TM) i3-4150 with 3.56Hz CPU and 4GB memory. The average run time (s) of the competing methods on the 12 test images (size: 256$\times$256) is shown in Table~\ref{lab:6}. Denoising 12 test images, NCSR, GID, LINC, MS-EPLL, AST-NLS and WNNM take, on average, roughly 348s, 346s, 257s, 191s, 459s and 202s, respectively.  For the test images, the proposed GSRC requires only 72s and 148s on average based pre-filtering BM3D and EPLL, respectively. Obviously, it can be seen that that the proposed GSRC  used less computation time than  these representative methods. Note that the run time of the proposed GSRC includes the pre-filtering process.
\section{Conclusion}
\label{6}
In this paper, we proposed a novel prior model named group sparsity residual constraint (GSRC) that exploited two kinds of nonlocal self-similar (NSS) prior and explored its application into image denoising. To boost the performance of group sparse-based image denoising, the  group sparsity residual was proposed, which is defined as the difference between the group sparse code of noisy  image and the group sparse code of the original image. Therefore, the problem of image denoising was translated into one that reduces the group sparsity residual. Since the original image was  unknown, to reduce the group sparsity residual, we first obtained some good estimation of the group sparse coefficients of the original image by pre-filtering and then the group sparse coefficients of the noisy  image were used to approximate the estimation. To enhance the accuracy of nonlocal similar patches selection, an adaptive patch search scheme was designed. In addition, to fuse these two NSS priors better, an iterative shrinkage algorithm was adopted to solve the GSRC model. Extensive experimental results have shown that the proposed GSRC can not only leads to visible PSNR improvements over many state-of-the-art methods such as BM3D and WNNM, but also preserves the image local structures, suppresses undesirable artifacts and results in a competitive speed. In the future, we will extend the proposed GSRC to other applications such as image deblurring, image super-resolution and image deblocking.

{\footnotesize

\end{document}